\documentclass[runningheads]{llncs}

\usepackage{eccv}

\usepackage{listings}
\usepackage{graphicx}
\usepackage{booktabs}
\usepackage{algorithm}
\usepackage{algorithmicx}
\usepackage{algpseudocode}

\usepackage{eccvabbrv}

\usepackage{graphicx}
\usepackage{booktabs}

\usepackage[accsupp]{axessibility}  

\usepackage{hyperref}

\usepackage{orcidlink}

\usepackage{multirow}
\usepackage{array}
\usepackage{tabularx}

\usepackage{makecell}

\usepackage{caption}

\usepackage{tikz}
\usetikzlibrary{tikzmark, calc}

\newcommand{\crosscell}[1]{%
\tikzmark{#1a}%
\tikzmark{#1b}%
}

\usetikzlibrary{shapes.arrows}

\usepackage{comment}
\usepackage{subcaption}
\usepackage{pifont}
\usepackage{bbold}
\usepackage{lipsum}

\usepackage{multibib}
\newcites{app}{Supplemental Material References}            
\begin{document}

\title{Neural Network Quantization by Learning Low-Loss Subspaces} 

\titlerunning{Neural Network Quantization by Learning Low-Loss Subspaces}

\author{Vladimir Protsenko\inst{ }\orcidlink{0000-0002-6191-7811} \and
Mikhalina Kharkevich\inst{ }\orcidlink{0009-0007-8208-2845}
\and
Alexander Vashchilko\inst{ }\orcidlink{0009-0004-6555-9013} 
\and
Vladimir Kryzhanovskiy\inst{ }\orcidlink{0009-0005-5339-8280}}

\authorrunning{V.~Protsenko~et al.}

\institute{Huawei, Montreal, Canada\\
\email{{\{protsenko.vladimir1, kryzhanovskiy.vladimir\}@huawei.com}}}

\maketitle

\begin{abstract}
Neural network quantization aims to find a discrete representation of parameters that preserves the performance of a full-precision (FP) model as faithfully as possible. Enforcing discrete constraints perturbs parameters away from a well-optimized minimum, generally resulting in performance degradation. Recent studies indicate that low-loss FP solutions are not isolated, but instead belong to connected low-loss subspaces of the loss landscape, where the loss maintains nearly the same minimum value. Models sampled from these subspaces are diverse and retain high accuracy. This raises the question: can a quantized model be constructed to lie within a low-loss subspace of the FP model, thereby automatically preserving performance? We address this question by learning quantization-aware linear paths in weight space optimized to minimize loss. We demonstrate that the midpoint of the resulting subspace is, by design, quantization-friendly and that its direct quantization yields performance comparable to that of quantization-aware training. The proposed procedure offers a novel perspective on weight quantization and, in contrast to conventional methods, neither relies on the straight-through estimator nor involves explicit discretization during training.
\end{abstract}
\section{Introduction}
\label{sec:intro}
Neural network quantization~\cite{nagel2021white,krishnamoorthi2018quantizing,gholami2022survey} has become one of the key methods for reducing computational cost and memory footprint of modern deep learning models. It aims to transform full-precision (FP) model parameters and inference operations into discrete representations, enabling efficient deployment on resource-constrained hardware platforms. Despite its practical advantages, quantization, particularly in the low-bit regime, remains challenging, since direct mapping of continuous FP parameters to a limited set of discrete values perturbs the optimized FP solution. These perturbations shift the model parameters away from well-optimized minima of the loss landscape~\cite{liu2021sharpness,draxler2018essentially}, often resulting in noticeable accuracy degradation.

Traditional quantization-aware training (QAT)~\cite{esser2019learned, javed2024qt, yu2024improving,nagel2021white,krishnamoorthi2018quantizing} and post-training quantization (PTQ)~\cite{frantar2022gptq,jacob2018quantization, nagel2019data, xiao2023smoothquant} methods seek to minimize the performance gap between FP and low-precision models by explicitly modeling discretization effects during optimization. In QAT pipelines, quantizers with non-differentiable rounding operations are embedded directly into the training graph to simulate low-precision behavior, while optimization relies on straight-through estimator (STE)–based~\cite{bengio2013estimating,yin2019understanding} surrogate gradients to enable backpropagation through discrete operators. Although STE-based training achieves strong empirical performance, its approximate gradients can introduce biased updates and unstable convergence~~\cite{liu2023bridging, pei2023quantization, lee2021network, nagel2022overcoming, ichikawa2025high, spallanzani2022training, yin2019understanding}, especially in low-bit regimes. Consequently, QAT methods are often highly sensitive to initialization~\cite{yun2025starting} and typically require starting from a well-trained FP model, making training quantized networks from scratch challenging. Moreover, the resulting models are usually optimized for a specific bit-width configuration, and changes in precision levels or quantization parameters frequently lead to accuracy degradation~\cite{chmiel2020robust, nahshan2021loss}. These limitations motivate the development of alternative quantization paradigms~\cite{chen2019metaquant, louizos2018relaxed, kim2021distance} capable of producing high-quality quantized solutions without using explicit discretization operators or relying on surrogate-gradient approximations.

Recent studies have demonstrated that well-converged deep neural networks are rarely confined to a single isolated minimum of a loss function~\cite{freeman2016topology, garipov2018loss, draxler2018essentially,ferbach2024proving}. Instead, their weights reside within extended low-loss subspaces~\cite{garipov2018loss, draxler2018essentially, frankle2020revisiting, ferbach2024proving} -- continuous regions of parameter space in which diverse solutions achieve nearly identical performance. These subspaces form smooth, connected manifolds that allow substantial variation in weights along specific directions while preserving low loss and high accuracy. In this context, studies on mode connectivity~\cite{draxler2018essentially, frankle2020linear, garipov2018loss} show that independently trained networks can be connected through simple linear or curved paths that remain entirely within low-loss regions. Furthermore, these continuous low-loss subspaces can be learned within a single training run~\cite{wortsman2021learning}, eliminating the need for multiple independent training procedures.

Motivated by these insights, we introduce a novel quantization approach that directly samples quantization-friendly solutions from the low-loss subspace of the FP model. During training, we learn a quantization-aware subspace that is explicitly constrained to remain within a region of low loss and high accuracy. Importantly, the proposed training procedure does not rely on straight-through estimator (STE) approximations and does not incorporate any discretization (\eg, rounding) operations during the training phase. Due to the design of the training objective, direct quantization of the midpoint of the learned linear subspace yields a low-precision model that remains within the same low-loss region, resulting in performance that closely matches its FP counterpart. Moreover, in many cases, the midpoint of the subspace achieves higher accuracy than the solution obtained via standard FP training, highlighting that subspace training induces an implicit regularization effect and improves generalization.

The main contributions of this work are threefold:

(i) \textbf{Subspace-based reformulation of quantization.} We reinterpret neural network quantization from the perspective of low-loss subspace optimization, moving beyond conventional discrete optimization with STE approximations. To this end, we propose the \textbf{Quantization by Learning Subspaces (QLS)} algorithm that learns quantization-aware linear paths in the full-precision parameter space and explicitly constructs solutions that are inherently robust to quantization. We demonstrate that QLS achieves performance comparable to, and in most cases exceeding, that of conventional quantization-aware training (QAT).

(ii) \textbf{Cross-bit adaptability.} By construction, models quantized with QLS can be directly deployed at higher bit-widths without retraining, consistently yielding monotonic improvements in accuracy. In contrast to conventional QAT, QLS solutions exhibit substantially reduced sensitivity to changes in bit-width and maintain stable performance when quantization scales are constrained to power-of-two values~\cite{li2019additive, yao2022rapq}.

(iii) \textbf{Strong initialization for QAT.} We show that the subspace-derived solution produced by QLS provides a strong initialization for subsequent QAT, resulting in improved convergence stability and higher final accuracy compared to standard full-precision initialization.

We note that the present study primarily focuses on weight-only quantization. This setting enables a direct validation of the proposed method and clearly illustrates the underlying principle as weight quantization can be naturally addressed by learning a simple linear subspaces in the parameter space. Conceptually, a learned subspace in the weight space induces structured variations in the corresponding activation space, introducing extra degrees of freedom, that can be leveraged to shape the activation distribution toward quantization-friendly representation. Extending the QLS method to joint weight–activation quantization is feasible and will be the subject of a separate upcoming paper. Here, we empirically demonstrate that the weight-quantized solution obtained with the same training budget as full-precision training provides a strong initialization for subsequent QAT of fully quantized models.

\section{Related Work}
\label{sec:previous_study}
\textbf{Loss Landscape Geometry and Mode Connectivity.} As modern neural networks involve increasing degrees of freedom, understanding the geometry of the loss landscape, particularly from the perspective of mode connectivity, has become critical, prompting extensive research in this direction~\cite{goodfellow2014qualitatively, ferbach2024proving, freeman2016topology, draxler2018essentially, garipov2018loss, wortsman2021learning, frankle2020revisiting}. Draxler~\etal~\cite{draxler2018essentially} showed that two independently trained low-loss solutions can often be connected by a continuous curve in weight space along which the training loss remains low, even when simple linear interpolation between the solutions exhibits a barrier. Garipov~\etal~\cite{garipov2018loss} proposed explicitly learning such low-loss connecting curves between two independently trained models. Wortsman~\etal~\cite{wortsman2021learning} generalized this idea to learned low-loss subspaces of various geometries. Instead of requiring multiple independently trained endpoints, Ref.~\cite{wortsman2021learning} shows that a low-loss subspace can be learned within a single training run. Frankle~\etal~\cite{frankle2020linear} demonstrated that linearly connected solutions are closely related to the lottery ticket phenomenon. An important observation of these studies is that central or averaged points within such low-loss regions often exhibit improved generalization and performance compared to standard (non-subspace) training. Stochastic Weight Averaging (SWA)~\cite{izmailov2018averaging} and followed weight-averaging strategies~\cite{wortsman2022model, zimmer2023sparse, li2022trainable, von2020neural} leverage this insight by averaging weights along the late-phase training trajectory to move the model toward flatter regions of the loss landscape. Foret~\etal~\cite{foret2020sharpness} introduced Sharpness-Aware Minimization (SAM), which jointly minimizes the training loss and its local sharpness by seeking parameters within a low-loss neighborhood, thereby improving generalization performance.

\textbf{Loss Landscape Geometry and Quantization.} Quantization robustness is closely associated with the geometric structure of the loss landscape~\cite{xia2025efficiently, liu2021sharpness, askarihemmat2024qgen, baldi2025loss, zhang2022quantization}. Xia~\etal~\cite{xia2025efficiently} propose a training strategy that explicitly moves optimization toward flat minima prior to quantizing by PTQ, arguing that reduced curvature around a converged FP solution reduces PTQ error. Liu~\etal~\cite{liu2021sharpness} extend sharpness-aware optimization to low-bit settings by modeling quantization noise as weight perturbations and optimizing the loss at perturbed weights to encourage convergence to flatter, quantization-robust minima. Similarly, Wang~\etal~\cite{wang2022squat} combine sharpness-aware updates with quantization-aware training in transformer architectures, while Jiang~\etal~\cite{jiang2025quantization} and Ma~\etal~\cite{ma2025learning} further explore flatness-oriented QAT strategies for improved out-of-distribution robustness and mixed-precision optimization. Shin~\etal~\cite{shin2020hlhlp} proposed a quantization training scheme that alternates between high and low precision training stages with a dynamically adjusted learning rate to guide convergence toward flatter minima in the loss surface. Overall, these studies indicate that shaping the local curvature properties near the solution improves robustness to quantization.

Despite methodological differences, these approaches remain local, meaning that they modify the optimization objective to bias a single solution toward a region of reduced sharpness, with the expectation that PTQ- or QAT-based quantization will incur lower accuracy degradation. At the intersection of subspace learning and quantization, Nunez~\etal~\cite{nunez2023lcs} propose the Learning Compressible Subspaces training, in which a low-dimensional compressible subspace is optimized to support multiple quantization levels, thereby enabling a controllable accuracy–efficiency trade-off at inference time. However, to the best of our understanding, the resulting ensemble of models is obtained by incorporating affine quantization~\cite{jacob2018quantization, nagel2021white} during training, and thus rely on quantization-aware optimization. To the best of our knowledge, all prior methods so far do not explicitly construct an extended low-loss subspace that is directly tailored for compatibility with a given discretization scheme.

\section{Approach}
\label{sec:approach}
\subsection{Quantization Preliminaries}
The $b$-bit uniform fake quantization operator $\hat{Q}_b(\cdot)$ can be written as~\cite{nagel2021white, esser2019learned}
\begin{equation}
\Theta^{({\rm quant})} = \hat{Q}_b(\Theta; s) = s\times\operatorname{clip}\left(\left\lfloor {\Theta}/{s} \right\rceil; B_{\rm b}^{\downarrow}, B_{\rm b}^{\uparrow}\right),
\label{quant_main_eq}
\end{equation}
where $\Theta$ denotes a full-precision weight or activation, $\Theta^{({\rm quant})}$ and $s^{-1}\times\Theta^{({\rm quant})}$ are the reconstructed values of $\Theta$ and its integer scaled representation, respectively. Here, $\operatorname{clip}(\Theta; B^{\downarrow}, B^{\uparrow})$ constrains $\Theta$ to $[B^{\downarrow}, B^{\uparrow}]$, and $\lfloor \Theta \rceil$ denotes rounding to the nearest integer. Throughout this work, we use the symmetric integer range $B_{\rm b}^{\uparrow(\downarrow)}=\pm (2^{b-1}-1)$ for signed data and $(B_{\rm b}^{\downarrow}, B_{\rm b}^{\uparrow}) = (0,\, 2^{b}-1)$ for unsigned data.

The scale factor $s$ in Eq.~(\ref{quant_main_eq}) determines the resolution of quantization grid. During QAT, it is typically treated as a learnable parameter (\eg, LSQ~\cite{esser2019learned} approach). A common initialization strategy for the weight scale factor is abs–max calibration~\cite{gholami2022survey}, defined as     
\begin{equation}
s^{\rm am}\left(\Theta\right)=\operatorname{ScaleByAbsMax}\left(\Theta; b\right)\equiv 2\max{\left(|\Theta|\right)}/(B_{\rm b}^{\uparrow}-B_{\rm b}^{\downarrow}).
\label{weight_calib}
\end{equation}

In practice, direct use of $s^{\rm am}$ given by Eq.~(\ref{weight_calib}) to quantize FP model, without an additional training process, typically leads to substantial quantization errors. In contrast, we demonstrate that our quantization-aware subspace training enables the construction of an FP solution whose high-quality quantized counterpart can be obtained directly in a post-training manner by applying Eqs.~(\ref{quant_main_eq}) and (\ref{weight_calib}).

\subsection{Quantization by Learning Subspaces (QLS)}
{\bf Quantization-Aware Subspace Training.} Let us consider weight quantization and denote by $\mathbf{\Theta} = (\Theta_{1}, \dots, \Theta_{N})$ the weights of the $N$ layers to be quantized. Here $\Theta_{p} \in \mathbb{R}^{c_{\mathrm{in},p} \times c_{\mathrm{out},p} \times k_{p} \times k_{p}}$ if $p$-th layer is a convolutional layer and $\Theta_{p} \in \mathbb{R}^{c_{\mathrm{in},p} \times c_{\mathrm{out},p}}$ in case of a linear layer. Let $\hat{Q}(\mathbf{\Theta}; \mathbf{s}) 
\equiv 
\big(
\hat{Q}_{b}(\Theta_{1}; s_{1}), \dots, \hat{Q}_{b}(\Theta_{N}; s_{N})
\big)$
denotes the quantized weights, where $\mathbf{s} = (s_{1}, \dots, s_{N})$ are per-layer scale factors. 
Without loss of generality, we assume all layers are quantized to the same bitwidth $b$ and apply per-tensor quantization, i.e., $s_{p} \in \mathbb{R}$.

Standard QAT minimizes the task loss $\mathcal{L}_{\mathrm{task}}$ under simulated quantization 
$\mathbf{\Theta}\rightarrow\hat{Q}(\mathbf{\Theta}; \mathbf{s}).
$
For weights this geometrically corresponds to optimizing a single point 
$\mathbf{\Theta} \in \mathbb{R}^{D}$ 
in a loss landscape of dimension 
$D$. Note that the non-quantized parameters are also optimized during training; however, for brevity, we omit them from the notation here and throughout the paper.

Instead of optimizing a single point using the STE approximation, we learn a linear low-loss subspace for each weight in a single training run. To this end, following the subspace learning algorithm proposed by Wortsman~\etal~\cite{wortsman2021learning}, the weights are parameterized as
\begin{equation}
\mathbf{\Theta}\rightarrow\mathbf{\Theta}^{(\pmb{\alpha})}
=
(\mathbf{J} - \mathbf{\pmb{\alpha}}) \odot \mathbf{\Theta}^{(1)}
+
\pmb{\alpha} \odot \mathbf{\Theta}^{(2)},
\end{equation}
where $\mathbf{\Theta}^{(1)}, \mathbf{\Theta}^{(2)} \in \mathbb{R}^{D}$ are trainable endpoints, 
$\pmb{\alpha}=(\alpha_{1},\dots,\alpha_{N})\sim \mathcal{U}([0,1]^{D})$ are independently sampled from a uniform distribution for each mini-batch, $\mathbf{J}\in\mathbb{R}^{D}$ is the all-ones tensor, and $\odot$ denotes element-wise multiplication. We note that, by \textbf{"linear subspace"} we refer to linearity along each individual weight. The midpoint of the resulting subspace of FP models is given as 
\begin{equation}
\mathbf{\Theta}^{(\mathrm{mid})} \equiv \mathbf{\Theta}^{(\pmb{\alpha}=1/2)} = (\mathbf{\Theta}^{(1)} + \mathbf{\Theta}^{(2)})/2.
\end{equation}

Our key idea is to dynamically adjust the element-wise L1 distances between the endpoints,
$\mathbf{d}
=
\big|
\mathbf{\Theta}^{(1)} - \mathbf{\Theta}^{(2)}
\big|
\in
\mathbb{R}^{D}$,
so that the quantized midpoint $\mathbf{\Theta}^{(\mathrm{quant})}
=
\hat{Q}\big(
\mathbf{\Theta}^{(\mathrm{mid})}; 
\mathbf{s} = \mathbf{s}^{(\mathrm{mid})}
\big)$
remains within the learned FP low-loss subspace. Here $\mathbf{s}^{\mathrm{(mid)}}=(s^{\mathrm{(mid)}}_{1},\dots,s^{\mathrm{(mid)}}_{N})$, where $s^{\mathrm{(mid)}}_{1\dots N}\in \mathbb{R}$ are given by

\begin{equation}
s^{\rm (mid)}_{p} = s^{\rm am}_{p}\left(\Theta^{(\rm mid)}_{p}\right)=\operatorname{ScaleByAbsMax}\left(\Theta_{p}^{(\rm mid)}; b\right).
\label{scale_mid}
\end{equation}

For uniform quantization of layer $p$ with step size $s^{\mathrm{(mid)}}_{p}$, the rounding error
\begin{equation}
\mathcal{E}_{p}
=s^{\mathrm{(mid)}}_{p}
\left\lfloor
{\Theta_{p}^{(\mathrm{mid})}}/{s^{\mathrm{(mid)}}_{p}}
\right\rceil
-
{\Theta_{p}^{(\mathrm{mid})}}
\end{equation}
lies strictly in the interval $s^{\mathrm{(mid)}}_{p}\times[-1/2, 1/2]$. 
Therefore, if for each layer $p$, the endpoint separation $d_{p}=\big|
\Theta^{(1)}_{p} - \Theta^{(2)}_{p}\big|$ satisfies 
\begin{equation}
\Delta_{p}=J_{p} - {d_{p}}/{s_{p}^{(\mathrm{mid})}} \leq 0 ,   
\label{distance_condition}
\end{equation}
where $J_{p}$ is the all-ones tensor of the same size as $d_{p}$, then both the floor and the ceil roundings of the midpoint remain within the learned low-loss linear paths consisting of FP models. 

To enforce this separation between the endpoints, we introduce the following regularization term into the training objective:
\begin{equation}
\mathcal{R}_{\rm qdist}\left(\mathbf{\Theta}^{(1), (2)}; b\right)=\frac{1}{N}\sum_{p=1}^{N}\frac{1}{D_p}\sum_{\text{all elements}}{
\mathcal{H}\left(\Delta_{p}\right)
\odot
\Delta_{p}^{2}},
\end{equation}
where $\mathcal{H}(\cdot)$ is the Heaviside step function and $D_p=\mathrm{dim}(\Theta_{p})$. $\mathcal{R}_{\rm qdist}$ penalizes deviations from the constraint defined by Eq.~(\ref{distance_condition}). $\mathcal{R}_{\rm qdist}\in[0, 1]$ and achieves $\mathcal{R}_{\rm qdist}=0$ when Eq.~(\ref{distance_condition}) is strictly satisfied. We evaluated several alternative forms of $\mathcal{R}_{\rm qdist}$, including linear and exponential variants, and found that the results are robust to its functional form (see Sec.~1 of the Supplemental Material).

The overall training objective can be formulated as
\begin{equation}
\min\limits_{\mathbf{\Theta}^{(1),(2)}}\Bigl(\mathbb{E}_{ (x, y) \sim \mathcal{D}_{\rm train}}\Bigl[\mathbb{E}_{\pmb{\alpha} \sim \mathcal{U}([0,1]^{D})} \Bigl[\mathcal{L}_{\rm task}\left(\mathbf{\Theta}^{(\pmb{\alpha})}\right)\Bigr]\Bigr] + \lambda_{\rm q}\mathcal{R}_{\rm qdist}\left(\mathbf{\Theta}^{(1),(2)};b\right)\Bigr),
\label{main_objective}
\end{equation}
where $\lambda_{\rm q}$ controls the strength of the regularization. For all the experiments, we set $\lambda_{\rm q}=1$.  

We highlight several key points of the proposed training procedure: (i) although the target bitwidth $b$ is included in training loop via Eq.~(\ref{scale_mid}), \textbf{no quantization is applied during training}; (ii) as \textbf{no rounding is involved}, optimization is performed entirely in FP with \textbf{no STE approximation}; (iii) the scale factors $\mathbf{s}^{\mathrm{(mid)}}$ are dynamically recomputed during each forward pass according to Eq.~(\ref{scale_mid}). In this sense, $\mathbf{s}^{\mathrm{(mid)}}$ can be interpreted as implicitly learnable parameters (but not independent parameters, as in the LSQ approach) that co-adapt with the endpoint distance to satisfy Eq.~(\ref{distance_condition}).

{\bf Quantization.} In our pipeline, quantization is performed post-training. After learning the low-loss linear subspace for each weight, the final quantized model is obtained by directly applying Eq.~(\ref{quant_main_eq}) to the midpoint $\mathbf{\Theta}^{(\mathrm{mid})}$ of the learned subspace, using the scale factor defined in Eq.~(\ref{scale_mid}). The overall QLS pipeline is summarized in Algorithm~\ref{alg:qls_algo} (see Sec.~2 of the Supplemental Material for the core pseudocode).

\textbf{By construction, rounding the midpoint yields a solution that corresponds to one of the FP models along the low-loss subspace.} Therefore, if the subspace consists entirely of low-loss, high-accuracy models (although in practice this assumption may not hold exactly; see Sec.~Results), quantization results in no loss of accuracy. Furthermore, since the scale factor is computed using the abs–max strategy, quantized weights for symmetric distributions are confined to the target integer range, thereby effectively eliminating clipping-induced errors.

Finally, the bitwidth during training and the bitwidth of the final quantized model are decoupled, meaning that specifying one for training does not require using the same for inference. To make this distinction explicit, we use the notation \textbf{QLS(x-bit train, y-bit inference)}, indicating that the subspace is learned to align with $x$-bit quantization, while the final model is quantized and evaluated at $y$-bit.
\begin{algorithm}[t!]
\caption{Quantization by Learning Low-Loss Subspaces (QLS)}
\label{alg:qls_algo}
\begin{algorithmic}[1]
\State{{\bfseries Input:} Network $f\left( \mathbf{\Theta}\right)$ of $N$ layers with weights $\mathbf{\Theta}=\left(\Theta_{1}, \dots, \Theta_{N}\right)$, train set $\mathcal{D}_{\rm train}$, bitwidth $b$, task loss function $\mathcal{L}_{\rm task}$, subspace quantization-aware regularization $\lambda_{q}\mathcal{R}_{\rm qdist}$. The weights are parameterized as $\mathbf{\Theta}\rightarrow\mathbf{\Theta}^{(\pmb{\alpha})}=\left(\mathbf{J}-\mathbf{\pmb{\alpha}}\right)\odot\mathbf{\Theta}^{(1)}+\pmb{\alpha}\odot\mathbf{\Theta}^{(2)}$, where $\mathbf{\Theta}^{(1, 2)}\in \mathbb{R}^{D}$ are trainable parameters, $\pmb{\alpha}\sim \mathcal{U}\left([0,1]^{D}\right)$.}
\State{{\bfseries Output:} Quantized weights $\mathbf{\Theta}^{({\rm quant})}$ are built based on a trained subspace.}
\State{\# \underline{Training of Quantization-Aware Subspaces}}
\For{$(x, y) \subseteq \mathcal{D}_{\rm train}$}
\For{each layer $p\in\left\{1\dots N\right\}$ in layers}
\State{Sample $\alpha_{p}$ of size $D_p=\mathrm{dim}(\Theta^{(1),(2)}_{p})$ from $\mathcal{U}([0,1]^{D_{p}})$}
\State $\Theta^{\left(\alpha\right)}_{p}=\left(J_{p}-\alpha_{p}\right)\odot\Theta_{p}^{(1)} + \alpha_{p}\odot\Theta_{p}^{(2)}$
\EndFor 
\State{$y^{\rm out} = f\left(\mathbf{\Theta}^{(\pmb{\alpha})}\right){\rm .forward}\left(x\right)$}
\State{$\mathcal{L} = \mathcal{L}_{\rm task}\left(y^{\rm out}, y\right) + \lambda_{\rm q}\mathcal{R}_{\rm qdist}\left(\mathbf{\Theta}^{({1}), ({2})};b\right)$}
\State{$\mathcal{L}{\rm .backward()}$}
\State{${\rm Update}\left(\mathbf{\Theta}^{(1)}, \mathbf{\Theta}^{(2)}\right)$}
\EndFor
\State{\# \underline{Weight Quantization (after train)}}
\For{each layer $p\in\left\{1\dots N\right\}$ in layers}
\State $\Theta^{({\rm mid})}_{p}=\left(\Theta_{p}^{(1)} + \Theta_{p}^{(2)}\right)/2$
\State $s^{\mathrm{(mid)}}_{p} = {\rm ScaleByAbsMax}\left(\Theta^{({\rm mid})}_{p}; b\right)$
\State $\Theta^{({\rm quant})}_{p} = \hat{Q}_b\left(\Theta^{({\rm mid})}_{p}; s^{\mathrm{(mid)}}_{p}\right)$
\EndFor
\State{$\mathbf{\Theta}^{({\rm quant})}=\left(\Theta^{({\rm quant})}_{1}, \dots, \Theta^{({\rm quant})}_{N}\right)$}
\end{algorithmic}
\end{algorithm}
\begin{figure}[t]  
    \includegraphics[width=1.0\linewidth]{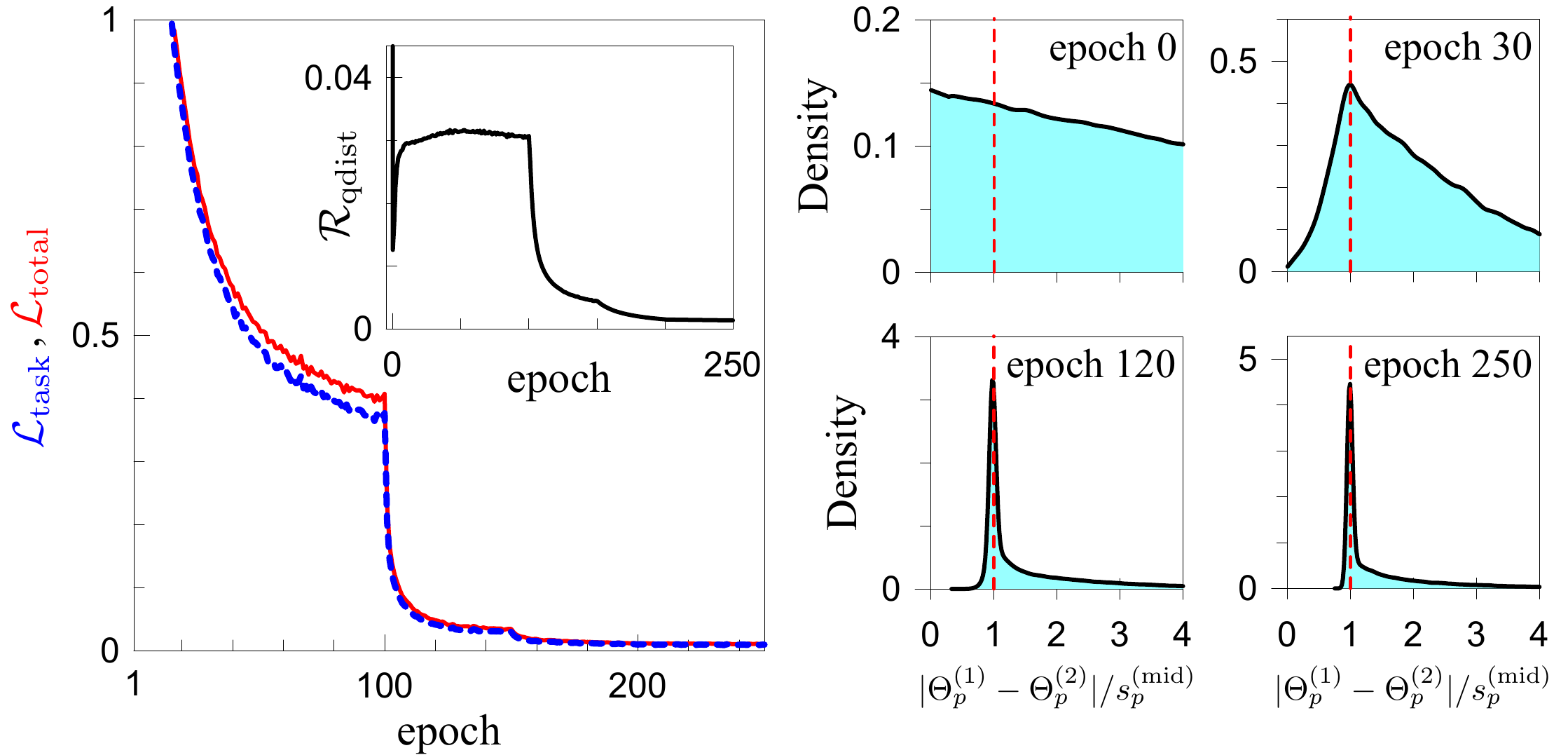}
    \caption{{\it Left panel:} The training loss functions $\mathcal{L}_{\rm task}$, $\mathcal{R}_{\rm qdist}\left(b=4\right)$ (inset) and $\mathcal{L}_{\rm total} = \mathcal{L}_{\rm task} + \lambda_{q}\mathcal{R}_{\rm qdist}$ for ResNet-$18$/CIFAR-$100$. {\it Right panels:} Density KDE-plots of the distribution of distances between the endpoints $d_{p} = |\Theta_{p}^{(1)} - \Theta_{p}^{(2)}|$ in the $p=\text{"layer3.0.conv2"}$ layer for different epochs ($d_{p}, \Theta_{p}^{(1),(2)} \in \mathbb{R}^{256 \times 256 \times 3 \times 3}$ and there are $D_{p}\approx 6\times 10^{5}$ values on the "x"-axis). The red vertical line shows distance $d_{p}=s_{p}^{(\mathrm{mid})}$.}
    \label{loss_distances}
\end{figure}
\section{Results}
\label{subsec:main_results}
In this section, we evaluate the proposed QLS algorithm under various settings. All layers except the first and the last are quantized. Implementation details and training protocols are provided in Secs.~2 and 3 of the Supplemental Material.
\subsection{Training of Quantization-Aware Low-Loss Subspaces}
\label{subsec:subspace_training}

Figure~\ref{loss_distances} illustrates the training loss (left) and the evolution of the distribution of distances  $\mathbf{d} = |\mathbf{\Theta}^{(1)} - \mathbf{\Theta}^{(2)}|$ (right). Initially (epoch 0), the endpoints are independently initialized, resulting in a broad distribution of distances. For each layer $p$, this distribution is characterized by a small fraction of distances below $s^{\mathrm{(mid)}}_{p}$ and a predominant contribution from distances exceeding $s^{\mathrm{(mid)}}_{p}$. Since the regularization $\mathcal{R}_{\rm qdist}$ penalizes only distances below $\mathbf{s}^{\mathrm{(mid)}}$, it induces only a minor perturbation to the task loss $\mathcal{L}_{\rm task}$ even at the beginning of training. 

As training progresses, the regularization suppresses distances below $\mathbf{s}^{\mathrm{(mid)}}$ while preventing previously distant endpoints from collapsing below this threshold. In later stages (e.g., epochs 120 and 250), the distance distribution becomes sharply concentrated around $\mathbf{d}=\mathbf{s}^{\mathrm{(mid)}}$, with a small residual tail for $\mathbf{d}>\mathbf{s}^{\mathrm{(mid)}}$, and $\mathcal{R}_{\rm qdist}$ becomes negligible. This training dynamic is consistent across experiments (\cref{tab:WOnlyClassificationFromFP,tab:WOnlyClassification,tab:WOnlySR}). Despite the regularization constraints, the task loss remains dominant optimization objective and converges to values comparable to or lower than those of standard training.

\begin{figure}[t]  
    \includegraphics[width=1.0\linewidth]{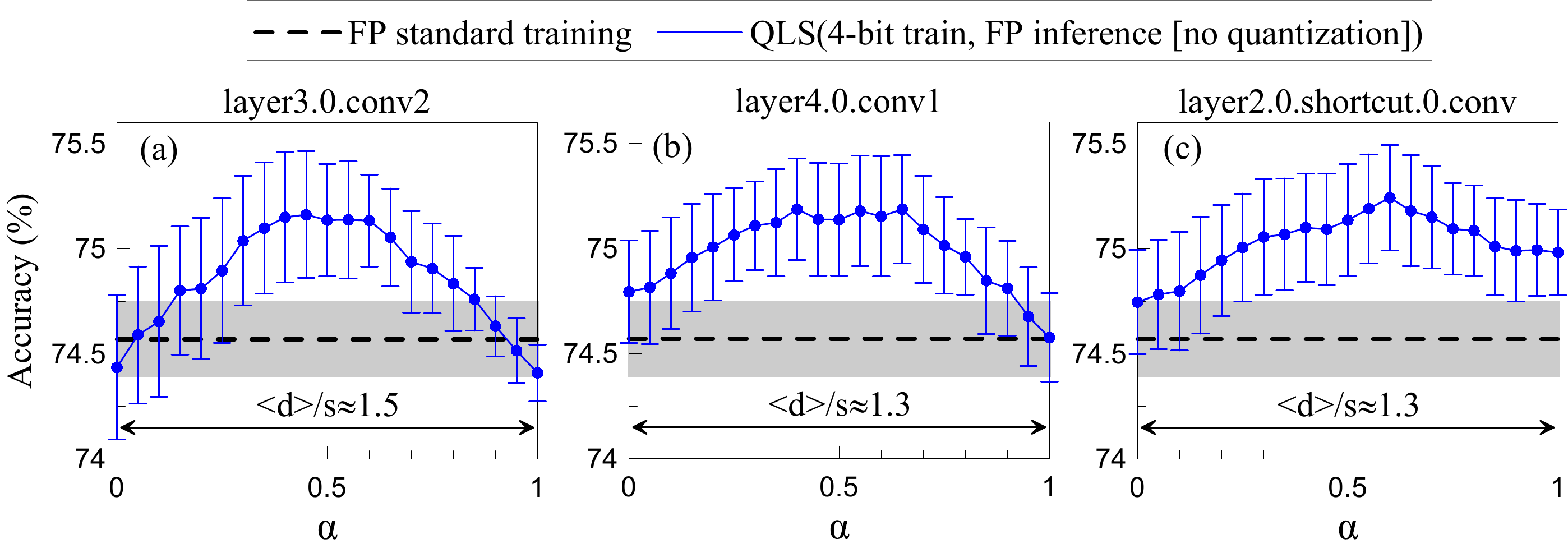}
    \caption{Test accuracy of ResNet-18 on CIFAR-100 as a function of $\alpha$ for a single selected layer (indicated in each panel). For that layer, $\Theta_{p}\rightarrow\left(1-\alpha\right)\Theta{p}^{(1)}+\alpha\Theta_{p}^{(2)}$, with $\alpha \in [0,1]$, while for all other layers $\alpha = 0.5$.}
    \label{acc_alpha_various_layers}
\end{figure}
To demonstrate that training yields a high-accuracy subspace between $\Theta^{(1)}_{p}$ $\left(\alpha_{p}=0\right)$  and $\Theta^{(2)}_{p}$ $\left(\alpha_{p}=1\right)$ for layer $p$, we evaluate model accuracy as a function of $\alpha_{p} = \alpha$ (Fig.~\ref{acc_alpha_various_layers}). For visualization, we constrain all elements within a layer to share a common scalar interpolation coefficient $\alpha \in \mathbb{R}$, although in general $\alpha_{p}$ has the same dimensionality as the corresponding weight tensor $\Theta^{(j)}_{p}, (j \in {1,2})$.

Despite this strong restriction on the degrees of freedom of $\alpha_{p}$, subspace training outperforms standard FP training over a substantial range of the interpolation region and, in some cases, across the entire subspace. In the majority of cases, the highest accuracy is achieved near the midpoint $\Theta^{({\rm mid})}_{p}$ (\ie for $\pmb{\alpha}=0.5$), in agreement with prior findings by Wortsman \etal~\cite{wortsman2021learning}. For each layer, the mean distance between endpoints (denoted in Fig.~\ref{acc_alpha_various_layers} by the double-arrow line) is on the order of the midpoint scale factor $\mathbf{s}^{(\mathrm{mid})}$ and is typically slightly larger due to the presence of distances exceeding this scale (see Fig.~\ref{loss_distances}).

In Tables~\ref{tab:WOnlyClassificationFromFP} and~\ref{tab:WOnlySR}, we compare FP models obtained via standard training (hereafter referred to as \textbf{"standard FP models"}) with the FP models corresponding to the midpoint of the learned subspace $\mathbf{\Theta}^{({\rm mid})}$ in QLS(4-, 5-, and 6-bit train, FP inference) settings. In almost all cases, the subspace-based FP model outperforms the counterpart trained with the same pipeline under standard training.

\begin{table}[t!]
	\caption{Average accuracy (\%) over 3 runs. For QLS, FP results correspond to QLS($b$-bit train, FP inference), while quantized results use matching training and inference bitwidths (i.e., QLS($b$-bit train, $b$-bit inference)). Best results are highlighted in red.}
	\label{tab:WOnlyClassificationFromFP}
	\centering
	\begin{tabular}{>{\centering}p{0.03\textwidth}>{\centering}p{0.00001\textwidth}>{\centering}p{0.01\textwidth}>{\centering}p{0.15\textwidth}>{\centering}p{0.2\textwidth}>{\centering}p{0.2\textwidth}>{\centering\arraybackslash}p{0.2\textwidth}}
		\Xhline{1.2pt}
	    \multicolumn{2}{l}{\multirow{2}{*}{Model \& Dataset}} & & \multicolumn{1}{l}{\multirow{2}{*}{FP training}} & \multicolumn{3}{c}{FP model and its direct quantization $\left(\rightarrow\right)$ by Eq.~(\ref{quant_main_eq})} \\ \cline{5-7}
        & & & & $\text{FP}\rightarrow \text{4-bit}$ & $\text{FP}\rightarrow \text{5-bit}$ & $\text{FP}\rightarrow \text{6-bit}$ \\ \hline
		\multirow{8}{*}{\rotatebox[origin=c]{90}{\parbox[c]{2.0cm}{\centering ResNet-$18$}}} & \multicolumn{1}{l}{\multirow{2}{*}{CIFAR-$10$}} & & \multicolumn{1}{l}{Standard} & \multicolumn{1}{c}{$93.75\rightarrow 92.53$} & \multicolumn{1}{c}{$93.75\rightarrow 93.46$} & \multicolumn{1}{c}{$93.75\rightarrow 93.69$}\\      
		& & & \multicolumn{1}{l}{QLS (Ours)} & \multicolumn{1}{c}{\textcolor{red}{$94.18 \rightarrow 94.07$}}  & \multicolumn{1}{c}{\textcolor{red}{$94.11 \rightarrow 94.1\phantom{0}$}} & \multicolumn{1}{c}{\textcolor{red}{$94.15 \rightarrow 94.15$}} \\ \cline{2-7}   
		& \multicolumn{1}{l}{\multirow{2}{*}{CIFAR-$100$}} & & \multicolumn{1}{l}{Standard} & \multicolumn{1}{c}{$74.57 \rightarrow 66.73$} & \multicolumn{1}{c}{$74.57 \rightarrow 73.49$} & \multicolumn{1}{c}{$74.57 \rightarrow 74.23$}\\     
		& & & \multicolumn{1}{l}{QLS (Ours)} & \multicolumn{1}{c}{\textcolor{red}{$75.14 \rightarrow 74.82$}}  & \multicolumn{1}{c}{\textcolor{red}{$75.13 \rightarrow 75.02$}} & \multicolumn{1}{c}{\textcolor{red}{$75.24 \rightarrow 75.12$}}\\ \cline{2-7} 
		& \multicolumn{1}{l}{\multirow{2}{*}{\begin{tabular}[l]{@{}l@{}}Tiny- \\ ImageNet\end{tabular}}} & & \multicolumn{1}{l}{Standard} & \multicolumn{1}{c}{$65.77 \rightarrow 38.21$} & \multicolumn{1}{c}{$65.77 \rightarrow 64.28$} & \multicolumn{1}{c}{$65.77 \rightarrow 65.21$} \\
		& & & \multicolumn{1}{l}{QLS (Ours)} & \multicolumn{1}{c}{\textcolor{red}{$66.11 \rightarrow 65.82$}}  & \multicolumn{1}{c}{\textcolor{red}{$66.43 \rightarrow 66.24$}} & \multicolumn{1}{c}{\textcolor{red}{$66.7\phantom{0} \rightarrow 66.7\phantom{0}$}} \\\cline{2-7}           
		& \multicolumn{1}{l}{\multirow{2}{*}{ImageNet}} & & \multicolumn{1}{l}{Standard} & \multicolumn{1}{c}{$\textcolor{red}{70.58} \rightarrow \phantom{0}1.79$} & \multicolumn{1}{c}{$70.58 \rightarrow 58.87$} & \multicolumn{1}{c}{$70.58 \rightarrow 69.04$} \\
		& & & \multicolumn{1}{l}{QLS (Ours)} & \multicolumn{1}{c}{$70.49 \rightarrow \textcolor{red}{69.64}$}  & \multicolumn{1}{c}{\textcolor{red}{$70.71 \rightarrow 70.46$}} & \multicolumn{1}{c}{\textcolor{red}{$71.1\phantom{0} \rightarrow 70.9\phantom{0}$}} \\ \hline
		\multirow{4}{*}{\rotatebox[origin=c]{90}{\parbox[c]{1.5cm}{\centering ResNet-$50$}}} & \multicolumn{1}{l}{\multirow{2}{*}{\begin{tabular}[l]{@{}l@{}}Tiny- \\ ImageNet\end{tabular}}} & & \multicolumn{1}{l}{Standard} & \multicolumn{1}{c}{$69.98 \rightarrow 35.94$} & \multicolumn{1}{c}{$69.98 \rightarrow 67.69$} & \multicolumn{1}{c}{$69.98 \rightarrow 69.18$}\\
		& & & \multicolumn{1}{l}{QLS (Ours)} & \multicolumn{1}{c}{$\textcolor{red}{70.1\phantom{0}} \rightarrow \textcolor{red}{70.01}$}  & \multicolumn{1}{c}{$\textcolor{red}{70.28} \rightarrow \textcolor{red}{70.14}$} & \multicolumn{1}{c}{$\textcolor{red}{70.31} \rightarrow \textcolor{red}{70.3\phantom{0}}$}\\ \cline{2-7}  
		& \multicolumn{1}{l}{\multirow{2}{*}{ImageNet}} & & \multicolumn{1}{l}{Standard} & \multicolumn{1}{c}{$\textcolor{red}{76.81} \rightarrow \phantom{0}0.13$} & \multicolumn{1}{c}{$\textcolor{red}{76.81} \rightarrow 54.64$} & \multicolumn{1}{c}{$76.81 \rightarrow 74.73$} \\
		& & & \multicolumn{1}{l}{QLS (Ours)} & \multicolumn{1}{c}{$76.43 \rightarrow \textcolor{red}{75.85}$}  & \multicolumn{1}{c}{$76.77 \rightarrow \textcolor{red}{76.56}$} & \multicolumn{1}{c}{$\textcolor{red}{76.85} \rightarrow \textcolor{red}{76.82}$}\\      
        \Xhline{1.2pt}
	\end{tabular}
\end{table}
\subsection{Weight Quantization in Linear Subspaces}
\label{subsec:quant_classification}
\subsubsection{QLS($b$-bit train, $b$-bit inference).} After quantization-aware subspace training, we quantize the midpoint $\mathbf{\Theta}^{({\rm mid})}$ by applying Eqs.~(\ref{quant_main_eq}) and (\ref{weight_calib}) with the same bitwidth $b$ used during training. As Table~\ref{tab:WOnlyClassificationFromFP} shows, $\mathbf{\Theta}^{({\rm mid})}$ is highly robust to quantization, unlike a standard FP model, which suffers significant degradation. This robustness is especially pronounced on challenging benchmarks (\eg, (Tiny-)ImageNet), where direct rounding of standard FP models severely degrades accuracy, whereas our method preserves performance. Notably, in some cases, selecting the subspace midpoint, which often surpasses the accuracy of standard training, allows models quantized by QLS to outperform their standard FP counterparts.

\begin{figure}[b!]  
    \includegraphics[width=1.0\linewidth]{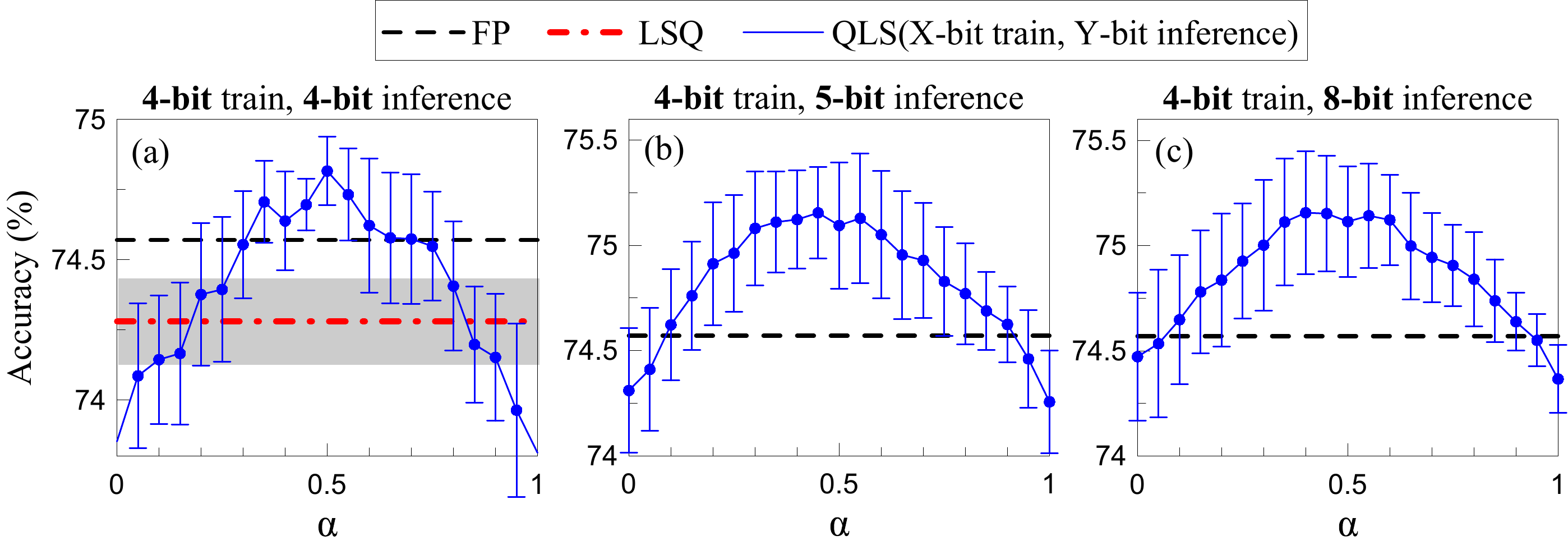}
    \caption{Test accuracy of QLS($4$-bit train, $4,5,6$-bit inference) quantized ResNet-$18$ on CIFAR-$100$ along the “layer3.0.conv2” subspace. All other layers use $\alpha = 0.5$.}
    \label{adaptation_layer_v2}
\end{figure}

We evaluate whether QLS can achieve the performance of STE-based QAT, where quantization is directly integrated into optimization. Using LSQ~\cite{esser2019learned} as a strong STE-based approach, we train models with both QLS and LSQ from scratch under the same protocol. Results on image classification (Table~\ref{tab:WOnlyClassification}) and super-resolution (Table~\ref{tab:WOnlySR}) tasks show that QLS consistently outperforms LSQ across nearly all settings.

Fig.~\ref{adaptation_layer_v2}~(a) shows accuracy when quantized weights are taken from subspace points other than the midpoint $\mathbf{\Theta}^{({\rm mid})}$. One can see that across a broad range of $\alpha_{p}=\alpha\in\mathbb{R}[0,1]$ around the midpoint, accuracy exceeds the results of LSQ (and even standard FP training).

\begin{table}[t!]
	\caption{Average accuracy (\%) over 3 runs. QLS uses matching training and inference bitwidths (i.e., QLS($b$-bit training, $b$-bit inference)). QLS(LSQ) $\rightarrow$ LSQ (1 epoch) denotes one additional LSQ epoch to QLS(LSQ) with $lr=10^{-7}$. FP $\rightarrow$ LSQ (1 epoch) denotes one LSQ epoch initialized from the standard FP baseline, \textbf{with the best learning rate} selected from $lr \in \{10^{-k}\}_{k=2}^{7}$. Best results are highlighted in red.}
	\label{tab:WOnlyClassification}
	\centering
	\begin{tabular}{>{\centering}p{0.05\textwidth}>{\centering}p{0.05\textwidth}>{\centering}p{0.01\textwidth}>{\centering}p{0.1\textwidth}>{\centering}p{0.15\textwidth}>{\centering}p{0.12\textwidth}>{\centering}p{0.12\textwidth}>{\centering\arraybackslash}p{0.12\textwidth}}
		\Xhline{1.2pt}
        \multicolumn{2}{l}{\multirow{2}{*}{\begin{tabular}[l]{@{}l@{}}Model \& Dataset \\ (FP baseline)\end{tabular}}} & & \multicolumn{1}{l}{\multirow{2}{*}{Epochs}} & \multicolumn{1}{l}{\multirow{2}{*}{Method}} & \multicolumn{3}{c}{Weight Bitwidth ($b$)} \\ \cline{6-8}        
        & & & & & 4 & 5 & 6 \\ \hline
		\multirow{10}{*}{\rotatebox[origin=c]{90}{\parbox[c]{2.0cm}{\centering ResNet-$18$}}} & \multicolumn{1}{l}{\multirow{2}{*}{\begin{tabular}[l]{@{}l@{}}CIFAR-$10$ \\ (93.75)\end{tabular}}} & & \multicolumn{1}{l}{\multirow{2}{*}{150}} & \multicolumn{1}{l}{LSQ~\cite{esser2019learned}} & \multicolumn{1}{c}{93.71} & \multicolumn{1}{c}{93.72} & \multicolumn{1}{c}{93.82}\\      
		& & & & \multicolumn{1}{l}{QLS (Ours)} & \multicolumn{1}{c}{\textcolor{red}{$94.07$}}  & \multicolumn{1}{c}{\textcolor{red}{$94.1\phantom{0}$}} & \multicolumn{1}{c}{\textcolor{red}{$94.15$}} \\ \cline{2-8}   
		& \multicolumn{1}{l}{\multirow{2}{*}{\begin{tabular}[l]{@{}l@{}}CIFAR-$100$ \\ (74.57)\end{tabular}}} & & \multicolumn{1}{l}{\multirow{2}{*}{250}} & \multicolumn{1}{l}{LSQ} & \multicolumn{1}{c}{74.28} & \multicolumn{1}{c}{74.25} & \multicolumn{1}{c}{74.57}\\     
		& & & & \multicolumn{1}{l}{QLS (Ours)} & \multicolumn{1}{c}{\textcolor{red}{$74.82$}}  & \multicolumn{1}{c}{\textcolor{red}{$75.02$}} & \multicolumn{1}{c}{\textcolor{red}{$75.12$}}\\ \cline{2-8} 
		& \multicolumn{1}{l}{\multirow{2}{*}{\begin{tabular}[l]{@{}l@{}}Tiny- \\ ImageNet\end{tabular}}} & & \multicolumn{1}{l}{\multirow{3}{*}{$100$}} & \multicolumn{1}{l}{\multirow{3}{*}{\begin{tabular}[l]{@{}l@{}}LSQ \\ QLS (Ours)\end{tabular}}} & \multicolumn{1}{c}{\multirow{3}{*}{\begin{tabular}[l]{@{}l@{}}65.65 \\ $\textcolor{red}{65.82}$\end{tabular}}} & \multicolumn{1}{c}{\multirow{3}{*}{\begin{tabular}[l]{@{}l@{}}65.8\phantom{0} \\$\textcolor{red}{66.24}$\end{tabular}}}& \multicolumn{1}{c}{\multirow{3}{*}{\begin{tabular}[l]{@{}l@{}}66.2\phantom{0} \\$\textcolor{red}{66.7\phantom{0}}$\end{tabular}}}\\
		& & & & & & & \multicolumn{1}{c}{} \\
		  & (65.77) & & & & & & \\\cline{2-8}
		& \multicolumn{1}{l}{\multirow{4}{*}{\begin{tabular}[l]{@{}l@{}}ImageNet \\ (70.58)\end{tabular}}} & & \multicolumn{1}{l}{\multirow{2}{*}{$100$}} & \multicolumn{1}{l}{LSQ} & \multicolumn{1}{c}{$\textcolor{red}{70.18}$} & \multicolumn{1}{c}{$70.37$} & \multicolumn{1}{c}{$70.6\phantom{0}$} \\
		& & & & \multicolumn{1}{l}{QLS (Ours)} & \multicolumn{1}{c}{$69.64$}  & \multicolumn{1}{c}{$\textcolor{red}{70.46}$} & \multicolumn{1}{c}{$\textcolor{red}{70.9\phantom{0}}$} \\ \cline{4-8}
        & & & \multicolumn{1}{l}{\multirow{2}{*}{$101$}} & \multicolumn{1}{l}{LSQ $\rightarrow$ LSQ(1 epoch)} & \multicolumn{1}{c}{$70.18$}  & \multicolumn{1}{c}{$70.38$} & \multicolumn{1}{c}{$70.58$} \\        
        & & & & \multicolumn{1}{l}{QLS $\rightarrow$ LSQ(1 epoch)} & \multicolumn{1}{c}{$\textcolor{red}{70.66}$}  & \multicolumn{1}{c}{$\textcolor{red}{70.7\phantom{0}}$} & \multicolumn{1}{c}{$\textcolor{red}{71.16}$} \\\hline
		\multirow{9}{*}{\rotatebox[origin=c]{90}{\parbox[c]{1.5cm}{\centering ResNet-$50$}}} & \multicolumn{1}{l}{\multirow{4}{*}{\begin{tabular}[l]{@{}l@{}l@{}}Tiny- \\ ImageNet \\ (69.98) \end{tabular}}} & & \multicolumn{1}{l}{\multirow{2}{*}{$100$}} & \multicolumn{1}{l}{LSQ} & \multicolumn{1}{c}{$67.47$} & \multicolumn{1}{c}{$69.14$} & \multicolumn{1}{c}{$70.09$}\\
		& & & & \multicolumn{1}{l}{QLS (Ours)} & \multicolumn{1}{c}{$\textcolor{red}{70.01}$}  & \multicolumn{1}{c}{$\textcolor{red}{70.14}$} & \multicolumn{1}{c}{$\textcolor{red}{70.3\phantom{0}}$}\\
        \cline{4-8}  
        & & & \multicolumn{1}{l}{\multirow{2}{*}{$101$}} & \multicolumn{1}{l}{LSQ $\rightarrow$ LSQ(1 epoch)} & \multicolumn{1}{c}{$67.37$}  & \multicolumn{1}{c}{$69.06$} & \multicolumn{1}{c}{$69.95$} \\
        & & & & \multicolumn{1}{l}{QLS $\rightarrow$ LSQ(1 epoch)} & \multicolumn{1}{c}{$\textcolor{red}{70.27}$}  & \multicolumn{1}{c}{$\textcolor{red}{70.26}$} & \multicolumn{1}{c}{$\textcolor{red}{70.28}$} \\\cline{2-8}     
		& \multicolumn{1}{l}{\multirow{5}{*}{\begin{tabular}[l]{@{}l@{}}ImageNet \\ (76.81) \end{tabular}}} & & \multicolumn{1}{l}{\multirow{2}{*}{$100$}} & \multicolumn{1}{l}{LSQ} & \multicolumn{1}{c}{$\textcolor{red}{76.1\phantom{0}}$} & \multicolumn{1}{c}{$76.41$} & \multicolumn{1}{c}{$76.68$} \\
		& & & & \multicolumn{1}{l}{QLS (Ours)} & \multicolumn{1}{c}{$75.85$}  & \multicolumn{1}{c}{$\textcolor{red}{76.56}$} & \multicolumn{1}{c}{$\textcolor{red}{76.82}$}\\
        \cline{4-8}  
        & & & \multicolumn{1}{l}{\multirow{3}{*}{$101$}} & \multicolumn{1}{l}{LSQ $\rightarrow$ LSQ(1 epoch)} & \multicolumn{1}{c}{$76.12$}  & \multicolumn{1}{c}{$76.33$} & \multicolumn{1}{c}{$76.66$} \\      
        & & & & \multicolumn{1}{l}{QLS $\rightarrow$ LSQ(1 epoch)} & \multicolumn{1}{c}{$\textcolor{red}{76.49}$}  & \multicolumn{1}{c}{$\textcolor{red}{76.77}$} & \multicolumn{1}{c}{$\textcolor{red}{76.87}$} \\
        & & & & \multicolumn{1}{l}{FP $\rightarrow$ LSQ(1 epoch)} & \multicolumn{1}{c}{$74.03$}  & \multicolumn{1}{c}{$75.83$} & \multicolumn{1}{c}{$76.03$} \\        
        \Xhline{1.2pt}
	\end{tabular}
\end{table}

We find that QLS is particularly effective when model capacity is comparable to or exceeds dataset size $(\text{capacity} \gtrsim \text{data})$. This is consistent with prior observations that such regimes exhibit geometrically extended low-loss regions~\cite{fan2025sharp, karhadkar2023mildly, wu2017towards, zhou2024md}, whereas data-rich settings $(\text{data} \gg \text{capacity})$ typically yield minima occupying smaller volumes in the loss landscape of smaller datasets~\cite{fan2025sharp}. In low-bit, data-rich scenarios, aligning the scale $\mathbf{s}^{\mathrm{(mid)}}$ with the underlying subspace geometry may require longer optimization compared to LSQ (\eg, 4-bit ImageNet). Nevertheless, with sufficient training, QLS outperforms both FP and LSQ baselines, even in these regimes (see Sec.~4 of the Supplemental Material).

These findings suggest that QLS can be particularly well suited for over- or balanced-parameterized modern transformer architectures. Recent studies~\cite{lee2025understanding, li2025seeking,li2024revisiting,zhoudoes} indicate that such models, including LLMs and diffusion models, exhibit pronounced flatness along multiple directions of the loss landscape, making them promising candidates for QLS; an analysis of this is left for future work.
\begin{table}[t]
	\caption{Average PSNR/SSIM for $3\times$ super-resolution with RFDN~\cite{liu2020residual} trained on DIV2K~\cite{agustsson2017ntire} and tested on Set5, Set14, and Urban100. Notation follows Table~\ref{tab:WOnlyClassification}.}
	\label{tab:WOnlySR}
	\centering
	\begin{tabular}{>{\centering}p{0.05\textwidth}>{\centering}p{0.03\textwidth}>{\centering}p{0.01\textwidth}>{\centering}p{0.15\textwidth}>{\centering}p{0.21\textwidth}>{\centering}p{0.21\textwidth}>{\centering\arraybackslash}p{0.21\textwidth}}
		\Xhline{1.2pt}
	    & \multicolumn{1}{l}{\multirow{2}{*}{bit}} & & \multicolumn{1}{l}{\multirow{2}{*}{Method}} & \multicolumn{3}{c}{Average PSNR/SSIM} \\ \cline{5-7}
        & & & & Set5  & Set14 &  Urban100\\\hline
        \multirow{11}{*}{\rotatebox[origin=c]{90}{\parbox[c]{3.5cm}{\centering RFDN (44 conv. layers)}}} & \multicolumn{1}{c}{\multirow{3}{*}{FP}} &  & Standard FP & \multicolumn{1}{c}{34.08 / 0.9237} & \multicolumn{1}{c}{30.17 / 0.838\phantom{0}} & \multicolumn{1}{c}{27.77 / 0.8425}\\
         &  &  & \multicolumn{1}{l}{QLS(3, FP)} & \multicolumn{1}{c}{34.1\phantom{0} / 0.9239} & \multicolumn{1}{c}{30.15 / 0.8379} & \multicolumn{1}{c}{27.78 / 0.8431}\\
         &  &  & \multicolumn{1}{l}{QLS(4, FP)} & \multicolumn{1}{c}{34.15 / 0.9242} & \multicolumn{1}{c}{30.18 / 0.8381} & \multicolumn{1}{c}{27.79 / 0.8434}\\\cline{2-7}
		& \multicolumn{1}{c}{\multirow{4}{*}{$3$}} &  & \multicolumn{1}{l}{LSQ} & \multicolumn{1}{c}{33.87 / 0.9219} & \multicolumn{1}{c}{30.00 / 0.8346} & \multicolumn{1}{c}{27.39 / 0.8344}\\
		&  &  & \multicolumn{1}{l}{QLS (Ours)} & \multicolumn{1}{c}{33.93 / 0.9225} & \multicolumn{1}{c}{30.01 / 0.8348} & \multicolumn{1}{c}{27.56 / 0.8382}\\ \cline{4-7} 
        &  &  & \multicolumn{1}{l}{LSQ $\rightarrow$ 1\% LSQ} & \multicolumn{1}{c}{33.88 / 0.9220} & \multicolumn{1}{c}{30.00 / 0.8347} & \multicolumn{1}{c}{27.42 / 0.8352}\\
        &  &  & \multicolumn{1}{l}{QLS $\rightarrow$ 1\% LSQ} & \multicolumn{1}{c}{\textcolor{red}{34.02 / 0.9233}} & \multicolumn{1}{c}{\textcolor{red}{30.08 / 0.8361}} & \multicolumn{1}{c}{\textcolor{red}{27.61 / 0.8391}}\\  \cline{2-7}
		& \multicolumn{1}{c}{\multirow{4}{*}{$4$}} &  & \multicolumn{1}{l}{LSQ} & \multicolumn{1}{c}{34.01 / 0.9230} & \multicolumn{1}{c}{30.10 / 0.8368} & \multicolumn{1}{c}{27.61 / 0.8396}\\
		&  &  & \multicolumn{1}{l}{QLS (Ours)} & \multicolumn{1}{c}{34.06 / 0.9237} & \multicolumn{1}{c}{30.09 / 0.8365} & \multicolumn{1}{c}{27.67 / 0.8408}\\ \cline{4-7} 
        &  &  & \multicolumn{1}{l}{LSQ $\rightarrow$ 1\% LSQ} & \multicolumn{1}{c}{34.01 / 0.9229} & \multicolumn{1}{c}{30.09 / 0.8363} & \multicolumn{1}{c}{27.61 / 0.8395}\\
        &  &  & \multicolumn{1}{l}{QLS $\rightarrow$ 1\% LSQ} & \multicolumn{1}{c}{\textcolor{red}{34.10 / 0.9239}} & \multicolumn{1}{c}{\textcolor{red}{30.11 / 0.8369}} & \multicolumn{1}{c}{\textcolor{red}{27.71 / 0.8414}}\\      
        \Xhline{1.2pt}
	\end{tabular}
\end{table}

As can be seen from Tables~\ref{tab:WOnlyClassificationFromFP} and~\ref{tab:WOnlySR}, the quantized midpoint, although located within the trained low-loss subspace, exhibits a slight decrease in accuracy compared to its FP counterpart. We attribute this behavior to the loss surface not being perfectly flat, with subtle local irregularities present. Thus, minor LSQ fine-tuning is expected to adjust the quantized weights away from high-loss local convexities and yield substantial performance improvement.

Overall, although QLS does not explicitly simulate quantization during optimization and follows the same training protocol as the FP baseline, it \textbf{simultaneously} achieves: (i) the FP model $\mathbf{\Theta}^{({\rm mid})}$ that generally outperforms the standard FP baseline, and (ii) the quantized model $\mathbf{\Theta}^{({\rm quant})} = \hat{Q}(\mathbf{\Theta}^{({\rm mid})}; \mathbf{s}^{\mathrm{(mid)}})$ that largely preserves accuracy of $\mathbf{\Theta}^{({\rm mid})} \gtrsim\text{(standard FP)}$ and, in most cases, outperforms STE-based optimization.
\begin{figure}[t]  
    \includegraphics[width=0.9\linewidth]{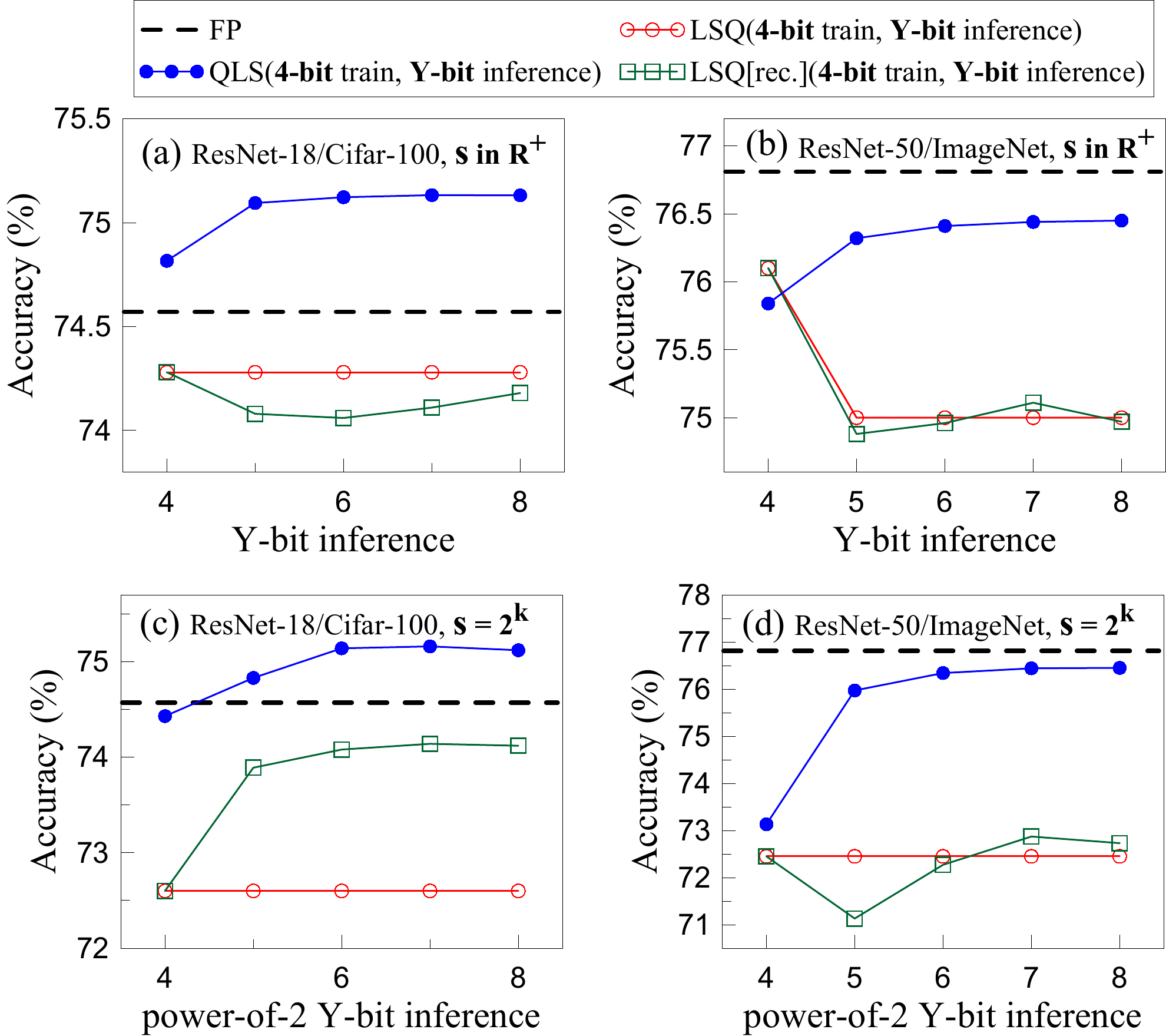}
    \caption{{\it Upper panels:} Test accuracy at higher bitwidths than training. {\it Lower panels:} As in Figs.~\ref{CrossAdaptation}a,b, with scale factors directly converted to the powers-of-two.}
    \label{CrossAdaptation}
\end{figure}
\subsubsection{LSQ Fine-Tuning.} To evaluate QLS under subsequent quantization-aware optimization, we perform brief LSQ fine-tuning with learning rate $lr \ll lr_{\rm start}$: one epoch for classification (Table~\ref{tab:WOnlyClassification}) and 1\% of full training for super-resolution (Table~\ref{tab:WOnlySR}). Fine-tuning of the converged LSQ solution produces only insignificant fluctuation-driven changes, while the QLS solution benefits from local weight adjustments, yielding significant accuracy gains. As a result, the (QLS $\rightarrow$ LSQ) pipeline consistently outperforms LSQ-based training alone. Notably, this performance cannot be achieved by LSQ, which starts from a standard FP model, regardless of the choice of $lr$ (see ResNet-50/ImageNet as a characteristic example), indicating that QLS provides a superior initialization for LSQ.

Finally, we find that QLS model weights are sparser than LSQ-based weights, and this persists after LSQ fine-tuning (see Sec.~5 of the Supplemental Material).

\subsubsection{Cross-Bit QLS($x$; $y\neq x$) and Power-of-Two Quantization.} Decoupling training $x$ and inference $y$ bitwidths in QLS allows direct transfer of a model to bitwidths different from those used in training. As shown in the upper panels of Fig.~\ref{CrossAdaptation}, QLS($x; y>x$) yields monotonic accuracy gains with increasing $y$.

This is a direct result of the subspace training: increasing the inference bitwidth $y$ induces an approximate power-of-two decay of the scale factor,
\begin{equation}
\mathbf{s}^{(\mathrm{mid})}(b=y) \approx 2^{x-y}\times\mathbf{s}^{(\mathrm{mid})}(b=x)\in \{ \tfrac{1}{2}, \tfrac{1}{4}, \tfrac{1}{8}, \dots \}\mathbf{s}^{(\mathrm{mid})}(b=x).
\end{equation}
Consequently, higher-bitwidth inference samples exponentially closer to the midpoint, progressively accessing higher-accuracy regions of the subspace (see Fig.~\ref{acc_alpha_various_layers}). Quantization of solutions beyond the midpoint also leads to progressive accuracy improvement (see Fig.~\ref{adaptation_layer_v2}) as $y$ increases, and the accuracy–$\alpha$ curve converges toward the FP subspace solution (Fig.~\ref{acc_alpha_various_layers}a).

In Fig.~\ref{CrossAdaptation}, we also evaluate adaptation to higher bitwidths for LSQ-based models using two conversion strategies. The first keeps scales obtained by LSQ, $\mathbf{s}(y)=\mathbf{s}(x)$, while expanding the integer range in Eq.~(\ref{quant_main_eq}). The second recalculates the scale factor according to the new granularity, $\mathbf{s}(y)=\mathbf{s}(x)\times({2^{x}-1})/({2^{y}-1})$ (see LSQ[rec.] in Fig.~\ref{CrossAdaptation}). In both cases, clipping/rounding errors lead to accuracy degradation or no improvement (when a model is robust to clipping errors).

In the lower panels of Fig.~\ref{CrossAdaptation}, we investigate the robustness of quantized models by directly converting scale factors to a hardware-friendly power-of-two representation without fine-tuning, $\mathbf{s} \rightarrow \mathbf{s}_{\rm pow2} = 2^{\lceil \log_2(\mathbf{s}) \rceil}$. Both QLS($x$; $x$) and QLS($x$; $y>x$) models outperform the corresponding LSQ-based models by a substantial margin.

For QLS($x$; $y<x$), the scale $\mathbf{s}$ extends beyond the trained subspace but has enough overlap to avoid severe accuracy loss, unlike LSQ-based models (see Sec.~6 of the Supplemental Material for additional results).

\begin{table}[t]
	\caption{Average PSNR of the W4A4 RFDN model (notation as in Table~\ref{tab:WOnlySR}). For (Standard FP $\rightarrow$ LSQ) the best result across $lr \in \{5\times 10^{-k}\}_{k=4}^{8}$ is presented.}
	\label{tab:W4A4}
	\centering
	\begin{tabular}{>{\centering}p{0.6\textwidth}>{\centering}p{0.01\textwidth}>{\centering}p{0.08\textwidth}>{\centering}p{0.12\textwidth}>{\centering\arraybackslash}p{0.12\textwidth}}
		\Xhline{1.2pt}
	   \multicolumn{1}{l}{Quantization Setup} & & Epochs & Set5 & Urban100\\\hline
        \multicolumn{1}{l}{(i) Standard FP $\rightarrow$ LSQ $\equiv$ $\text{W4}_{\text{LSQ}}\text{A4}_{\text{LSQ}}$} & & \multicolumn{1}{l}{1000} & 31.01  & 24.89 \\
        \multicolumn{1}{l}{(ii) $\mathbf{\Theta}^{({\rm mid})}$ $\rightarrow$ $\text{W4}_{\text{QLS (ours)}}\text{A4}_{\text{LSQ}}$} & & \multicolumn{1}{l}{10} &  32.16 & 26.01 \\
        \multicolumn{1}{l}{(iii) $\mathbf{\Theta}^{({\rm mid})}$ $\rightarrow$ $\text{W4}_{\text{QLS (ours)}}\text{A4}_{\text{LSQ}}$} & & \multicolumn{1}{l}{1000} & 32.85 & 26.29 \\      
        \Xhline{1.2pt}
	\end{tabular}
\end{table}
\subsubsection{Full Quantization.} Table 4 shows results for 4-bit weight and $4$-bit activation quantization (W4A4). We use the RFDN model~\cite{liu2020residual} as a representative image-to-image architecture, where low-bit activation quantization is particularly challenging due to poor convergence under STE-based optimization. Starting from $\mathbf{\Theta}^{(\mathrm{mid})}$ trained with $\mathcal{R}_{\rm qdist}(b=4)$ under the standard FP pipeline, we apply QLS to weights and LSQ to activations. This combined approach significantly outperforms the conventional FP $\rightarrow$ LSQ quantization, even with limited training (case (ii); see the Supplementary Material for additional results).  
\section{Conclusion} We propose QLS, a quantization strategy that constructs quantized weights directly from the linear low-loss subspaces of a full-precision model. QLS does not use rounding operations during training and does not rely on STE approximations. In the vast majority of experiments, the subspace midpoint simultaneously (i) achieves superior FP accuracy compared to standard training and (ii) produces a quantized solution that outperforms LSQ-based QAT.

Due to its subspace formulation, QLS exhibits strong bitwidth-agnostic behavior and robustness to power-of-two quantization. Furthermore, QLS provides an effective initialization for subsequent QAT training, particularly in low-bit joint weight-activation quantization scenarios where STE-based methods exhibit slow or suboptimal convergence (see discussion on limitations and future work in the Supplementary Material).

\bibliographystyle{splncs04}
\bibliography{main}

\begin{thebibliography}{10}
\providecommand{\url}[1]{\texttt{#1}}
\providecommand{\urlprefix}{URL }
\providecommand{\doi}[1]{https://doi.org/#1}

\bibitem{agustsson2017ntire_app}
Agustsson, E., Timofte, R.: Ntire 2017 challenge on single image super-resolution: Dataset and study. In: Proceedings of the IEEE conference on computer vision and pattern recognition workshops. pp. 126--135 (2017)

\bibitem{bevilacqua2012low}
Bevilacqua, M., Roumy, A., Guillemot, C., Alberi-Morel, M.L.: Low-complexity single-image super-resolution based on nonnegative neighbor embedding  (2012)

\bibitem{deng2009imagenet}
Deng, J., Dong, W., Socher, R., Li, L.J., Li, K., Fei-Fei, L.: Imagenet: A large-scale hierarchical image database. In: 2009 IEEE conference on computer vision and pattern recognition. pp. 248--255. Ieee (2009)

\bibitem{huang2015single}
Huang, J.B., Singh, A., Ahuja, N.: Single image super-resolution from transformed self-exemplars. In: Proceedings of the IEEE conference on computer vision and pattern recognition. pp. 5197--5206 (2015)

\bibitem{kingma2014adam}
Kingma, D.P., Ba, J.: Adam: A method for stochastic optimization. arXiv preprint arXiv:1412.6980  (2014)

\bibitem{krizhevsky2009learning}
Krizhevsky, A., Hinton, G., et~al.: Learning multiple layers of features from tiny images  (2009)

\bibitem{le2015tiny}
Le, Y., Yang, X., et~al.: Tiny imagenet visual recognition challenge. CS 231N  \textbf{7}(7), ~3 (2015)

\bibitem{liu2020residual_app}
Liu, J., Tang, J., Wu, G.: Residual feature distillation network for lightweight image super-resolution. In: Computer Vision--ECCV 2020 Workshops: Glasgow, UK, August 23--28, 2020, Proceedings, Part III 16. pp. 41--55. Springer (2020)

\bibitem{russakovsky2015imagenet}
Russakovsky, O., Deng, J., Su, H., Krause, J., Satheesh, S., Ma, S., Huang, Z., Karpathy, A., Khosla, A., Bernstein, M., et~al.: Imagenet large scale visual recognition challenge. International journal of computer vision  \textbf{115},  211--252 (2015)

\bibitem{wortsman2021learning_app}
Wortsman, M., Horton, M.C., Guestrin, C., Farhadi, A., Rastegari, M.: Learning neural network subspaces. In: International Conference on Machine Learning. pp. 11217--11227. PMLR (2021)

\bibitem{zeyde2012single}
Zeyde, R., Elad, M., Protter, M.: On single image scale-up using sparse-representations. In: Curves and Surfaces: 7th International Conference, Avignon, France, June 24-30, 2010, Revised Selected Papers 7. pp. 711--730. Springer (2012)

\end{thebibliography}

\clearpage

\title{Supplemental Material to ``Neural Network Quantization by Learning Low-Loss Subspaces''}

\titlerunning{Neural Network Quantization by Learning Low-Loss Subspaces}

\author{Vladimir Protsenko\inst{ }\orcidlink{0000-0002-6191-7811} \and
Mikhalina Kharkevich\inst{ }\orcidlink{0009-0007-8208-2845}
\and
Alexander Vashchilko\inst{ }\orcidlink{0009-0004-6555-9013} 
\and
Vladimir Kryzhanovskiy\inst{ }\orcidlink{0009-0005-5339-8280}}

\authorrunning{V.~Protsenko~et al.}

\institute{Huawei, Montreal, Canada\\
\email{{\{protsenko.vladimir1, kryzhanovskiy.vladimir\}@huawei.com}}}

\maketitle
\section{Effect of the Functional Form of $\mathcal{R}_{\rm qdist}$}
\label{sec:R_comparison}

In Eq.~(8) of the main text (see Sec.~3.2), we introduce the quantization-aware regularization term
\begin{equation}
\mathcal{R}_{\rm qdist}\left(\mathbf{\Theta}^{(1)}, \mathbf{\Theta}^{(2)}\right)=\frac{1}{N}\sum_{p=1}^{N}\frac{1}{D_p}\sum_{\text{all elements}}{F\left(\Delta_{p}\left(\Theta^{(1)}_{p}, \Theta^{(2)}_{p}\right)\right)},
\end{equation}
which penalizes violations of the constraint $\Delta_p\left(\Theta^{(1)}_{p}, \Theta^{(2)}_{p}\right) \leq 0$.

In the main paper, we use the quadratic form of the penalty function $F(\cdot)$
\begin{equation}
F^{(\mathrm{A})}(\Delta)=
\mathcal{H}(\Delta)\odot\Delta^2,
\end{equation}
where $\mathcal{H}(\cdot)$ denotes the element-wise Heaviside step function. This choice yields a penalty that grows quadratically when the constraint $\Delta_p \leq 0$ is violated.

The regularization principle underlying the QLS method is not tied to this particular analytic form. More generally, any monotone penalty applied to positive values of $\Delta_p$ enforces the same structural constraint. To investigate the sensitivity of the QLS method to the specific choice of $F(\cdot)$, we evaluate two alternative functional forms.

First, we consider a sigmoid-based penalty
\begin{equation}
F^{(\rm B)}(\Delta)=\sigma\left(10\Delta - 5\right),
\qquad
\sigma(x)=\left(1+e^{-x}\right)^{-1},
\label{sigmoid_based}
\end{equation}
which introduces a smooth exponentially decaying penalty.

Second, we consider a linear ReLU-based penalty
\begin{equation}
F^{(\mathrm{C})}(\Delta)=
\mathrm{ReLU}(\Delta)=\max(0,\Delta).
\end{equation}

\begin{figure}[t]
    \centering
    \includegraphics[width=0.50\linewidth]{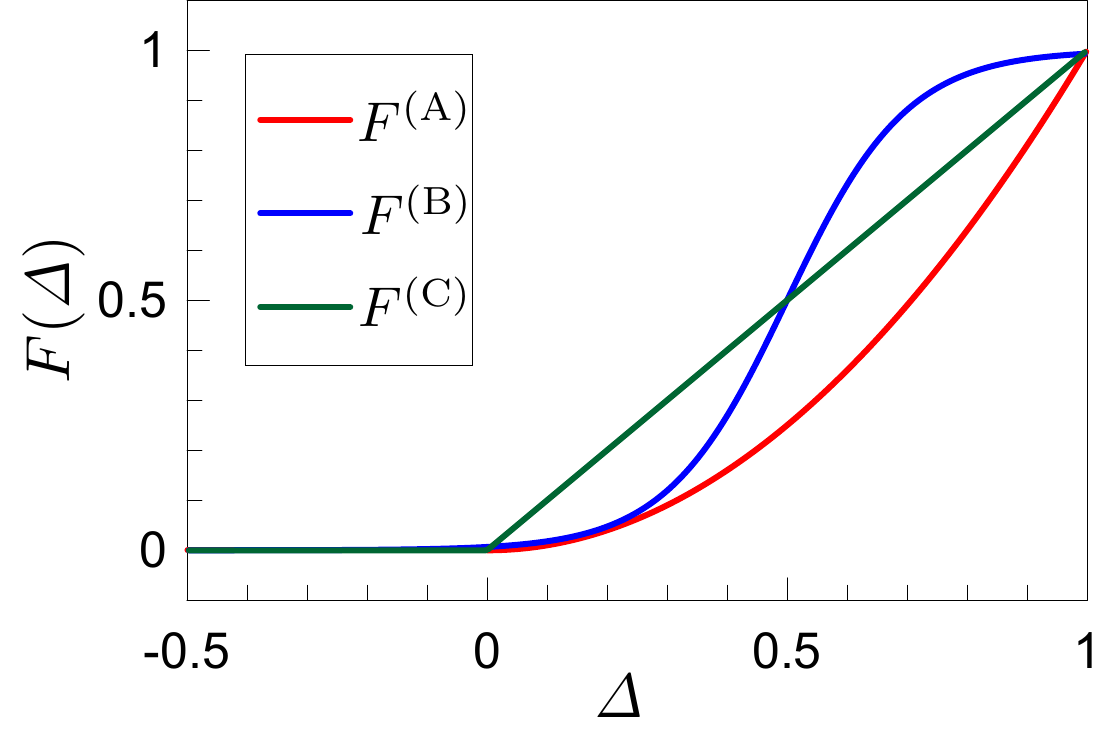}
    \caption{Functional forms of the penalty function  $F(\Delta)$.}
    \label{Regularization}
\end{figure}

These choices induce different curvature properties of the penalty: the quadratic form produces polynomial growth, the sigmoid introduces an exponential transition, and the ReLU yields a piecewise-linear penalty. Figure~\ref{Regularization} illustrates the resulting penalty profiles.

Table~\ref{tab:regularization_form} shows the accuracy obtained with the regularization forms introduced above. All experiments follow the same training protocol used in the main paper (see \cref{training_protocols} of this Supplemental Material); the only modification is the functional form of $F(\Delta)$. Across all experiments, the three variants yield comparable performance. We find that the sigmoid-based formulation $F(\Delta)=F^{(\rm B)}(\Delta)$ exhibits slightly larger variance across runs, which we attribute to slower optimization dynamics due to sigmoid saturation. Aside from this effect, the overall performance of the QLS method remains qualitatively the same.

Thus, the QLS gains are not tied to a particular analytic form of the regularization. Different penalty functions may nevertheless show different convergence behavior. In practice, we adopt the quadratic formulation $F^{(\mathrm{A})}$, which empirically yields faster convergence and slightly more stable training. Identifying an optimal penalty for enforcing quantization-aware constraints in subspace training remains an interesting direction for future work.
\begin{table}[t]
	\caption{Average accuracy (\%) over 3 runs of the QLS method for different functional forms of the penalty function $F(\Delta)$. QLS uses matching training and inference bitwidths (\ie, QLS($b$-bit training, $b$-bit inference)).}
	\label{tab:regularization_form}
	\centering
	\begin{tabular}{>{\centering}p{0.05\textwidth}>{\centering}p{0.02\textwidth}>{\centering}p{0.13\textwidth}>{\centering}p{0.13\textwidth}>{\centering}p{0.13\textwidth}>{\centering\arraybackslash}p{0.13\textwidth}}
		\Xhline{1.2pt}
        \multicolumn{1}{l}{\multirow{2}{*}{Model \& Dataset}} & & \multicolumn{1}{l}{\multirow{2}{*}{Penalty}} & \multicolumn{3}{c}{Weight Bitwidth ($b$)} \\ \cline{4-6}
        & & &  4 & 5 & 6 \\\hline
        & & & & & \\ [-0.35cm]
		\multicolumn{1}{l}{\multirow{3}{*}{\begin{tabular}[l]{@{}l@{}} ResNet-$18$\\ CIFAR-$100$\end{tabular}}} & &  \multicolumn{1}{l}{$F^{(\mathrm{A})}$} & \multicolumn{1}{c}{74.82} & \multicolumn{1}{c}{75.02} & \multicolumn{1}{c}{75.12}\\
		   & & \multicolumn{1}{l}{$F^{(\mathrm{B})}$} & \multicolumn{1}{c}{74.77} & \multicolumn{1}{c}{75.10} & \multicolumn{1}{c}{75.13}\\ 
		   &  & \multicolumn{1}{l}{$F^{(\mathrm{C})}$} & \multicolumn{1}{c}{74.71} & \multicolumn{1}{c}{75.17} & \multicolumn{1}{c}{75.25}\\\hline
        & & & & & \\ [-0.35cm]
		\multicolumn{1}{l}{\multirow{3}{*}{\begin{tabular}[l]{@{}l@{}} ResNet-$18$\\TinyImageNet\end{tabular}}} & &  \multicolumn{1}{l}{$F^{(\mathrm{A})}$} & \multicolumn{1}{c}{65.82} & \multicolumn{1}{c}{66.24} & \multicolumn{1}{c}{66.7\phantom{0}}\\
		   & & \multicolumn{1}{l}{$F^{(\mathrm{B})}$} & \multicolumn{1}{c}{65.78} & \multicolumn{1}{c}{66.01} & \multicolumn{1}{c}{66.53}\\ 
		   &  & \multicolumn{1}{l}{$F^{(\mathrm{C})}$} & \multicolumn{1}{c}{65.91} & \multicolumn{1}{c}{66.1\phantom{0}} & \multicolumn{1}{c}{66.56}\\           
        \Xhline{1.2pt}
	\end{tabular}
\end{table}

\section{Implementation Details and Pseudocode}
\label{sec:code}

In this section, we provide a more detailed description of the pseudocode of the core components of the proposed QLS pipeline, illustrated for convolutional layers for brevity. Integrating the method into standard training pipelines is straightforward and requires only minor modifications. Specifically, (i) convolutional layers that are to be quantized are replaced with subspace convolutional layers (\cref{alg:qlsconv}; the algorithms are provided at the end of this supplementary material); and (ii) the training loop (\cref{alg:train_classification}) is modified to include the sampling of weights from the subspace and to incorporate the quantization-aware regularization term (see~\cref{alg:qdist}) into the total training loss. In practice, the sampling procedure can be equivalently implemented within the forward pass of the subspace convolution layer when operating in training mode.

\section{Details of the Experiments}
\label{sec:training}
\subsection{Datasets}
For image classification, we conduct experiments on four datasets: CIFAR-$10$~\citeapp{krizhevsky2009learning}, CIFAR-$100$~\citeapp{krizhevsky2009learning}, TinyImageNet~\citeapp{le2015tiny}, and ImageNet (ILSVRC-2012)~\citeapp{russakovsky2015imagenet, deng2009imagenet}.

The CIFAR-10 and CIFAR-100 datasets each contain 60000 RGB images of spatial resolution $32\times32$, organized into 10 and 100 classes, respectively. Following common practice, we split each dataset into $45000$ training images, $5000$ validation images, and $10000$ test images. During training, we apply random horizontal flips for data augmentation.

The TinyImageNet dataset consists of 200 object classes with a total of $120000$ images. The dataset includes $100000$ training images, $10000$ validation images, and $10000$ test images, all with spatial resolution $64\times64$. During training, we apply random horizontal flips for data augmentation.

Finally, the ImageNet dataset contains approximately $1.28$ million training images and $50000$ validation images divided onto $1000$ categories. Following standard pipeline, images are resized to $256\times256$ pixels and augmented with random $224\times224$ crops and random horizontal flips during training.

For the $3\times$ image super-resolution task, we use the RFDN model~\citeapp{liu2020residual_app} and train it on the DIV2K~\citeapp{agustsson2017ntire_app} dataset. The dataset contains 900 high-resolution (2K resolution) images, split into 800 training images and 100 validation images. During training, we randomly crop patches of size $24\times24$ from the low-resolution inputs and the corresponding $72\times72$ patches from the high-resolution targets. Following the training protocol of RFDN~\citeapp{liu2020residual_app}, the dataset is further augmented using random horizontal flips and $90^\circ$ rotations. Test performance is evaluated using Peak Signal-to-Noise Ratio (PSNR), computed on the luminance (Y) channel of the YCbCr color space. We report results on the standard super-resolution benchmarks Set5, Set14, and Urban100, which contain 5, 14, and 100 images, respectively~\citeapp{bevilacqua2012low,zeyde2012single, huang2015single}.

\subsection{Training Protocols}
\label{training_protocols}
For all experiments, we use the original torchvision implementations of ResNet-$18$ and ResNet-$50$, applying dataset-specific architectural adaptations following the subspace training setups of Wortsman~\etal~\citeapp{wortsman2021learning_app}. The RFDN model is implemented according to the original paper~\citeapp{liu2020residual_app}.

Unless otherwise specified in the main text, models trained from scratch follow the protocols below:

{\bf ResNet-18 (CIFAR-10).} The model is trained for 150 epochs with an initial learning rate of $10^{-3}$. The learning rate is decayed by a factor of 10 at epochs 75, 100, and 125. We use a batch size of 64 and optimize with Adam~\citeapp{kingma2014adam} with a weight decay of $4\times 10^{-5}$.

{\bf ResNet-18 (CIFAR-100).} The model is trained for 250 epochs with an initial learning rate of $10^{-3}$, decayed by a factor of 10 at epochs 100, 150, and 200. The batch size is 64, and Adam is used with the same weight decay as above.

{\bf ResNet-18 \& ResNet-50 (TinyImageNet).} Models are trained for 300 epochs with a batch size of 128. The initial learning rate is 0.1 and is decayed to zero using a cosine scheduler. Optimization is performed using SGD with a weight decay of $5\times10^{-4}$.

{\bf ResNet-18 \& ResNet-50 (ImageNet).} Models are trained for 100 epochs with a batch size of 256. The initial learning rate is 0.1 and decayed to zero via a cosine scheduler. Optimization is performed with SGD and a weight decay of~$1\times10^{-4}$.

{\bf RFDN (DIV2K).} The model is trained for 1000 epochs with an initial learning rate of $5\times10^{-4}$, decayed to zero via a cosine scheduler. The batch size is 32, and the model is optimized using Adam without weight decay.
\begin{table}[b!]
	\caption{Average accuracy (\%) over 3 runs. The QLS results correspond to QLS($4$-bit train, $4$-bit inference). All other notation and parameters are as in Table~2 of the main text.}
	\label{tab:additional_train}
	\centering
	\begin{tabular}{>{\centering}p{0.05\textwidth}>{\centering}p{0.05\textwidth}>{\centering}p{0.14\textwidth}>{\centering}p{0.14\textwidth}>{\centering\arraybackslash}p{0.14\textwidth}}
		\Xhline{1.2pt}
        \multicolumn{1}{l}{\multirow{2}{*}{Model \& Dataset}} & & \multicolumn{1}{l}{\multirow{2}{*}{Epochs}} & \multicolumn{2}{c}{Method} \\ \cline{4-5}
        & & & LSQ & QLS (Ours) \\\hline
		\multicolumn{1}{l}{\multirow{2}{*}{\begin{tabular}[l]{@{}l@{}} W4Afp ResNet-$18$\\ ImageNet\end{tabular}}} & &  \multicolumn{1}{l}{100}& \multicolumn{1}{c}{70.18} & \multicolumn{1}{c}{69.64}\\
		   & & \multicolumn{1}{l}{200} & \multicolumn{1}{c}{70.61} & \multicolumn{1}{c}{70.75}\\\hline
		\multicolumn{1}{l}{\multirow{2}{*}{\begin{tabular}[l]{@{}l@{}} W4Afp ResNet-$50$\\ ImageNet\end{tabular}}} & &  \multicolumn{1}{l}{100}& \multicolumn{1}{c}{76.1\phantom{0}} & \multicolumn{1}{c}{75.85}\\
		   & & \multicolumn{1}{l}{200} & \multicolumn{1}{c}{76.53} & \multicolumn{1}{c}{76.64}\\
        \Xhline{1.2pt}
	\end{tabular}
\end{table}
\section{Effect of Extended Training in Low-Bit Regimes}
\label{sec:lowbit_ImageNet}
As reported in the main paper, the QLS quantization consistently outperforms LSQ across the evaluated settings, with the sole exception of 4-bit ResNet-family models trained on the ImageNet dataset. As discussed in the main text, we attribute this behavior to the slightly slower convergence of QLS compared to LSQ, which explicitly incorporates quantizers into the optimization process. This effect becomes pronounced in low-bit, data-rich regimes.

In the experiments reported in the main paper, both LSQ and QLS are trained for the same number of epochs as the corresponding full-precision models, namely 100 epochs. Under this fixed training budget, the faster convergence of LSQ leads to slightly better performance in this particular setting. However, when the training horizon is extended (see Table~\ref{tab:additional_train}), we observe that QLS eventually matches or surpasses LSQ even in this regime.

\section{Weight Sparsity}
\label{sec:sparsity}
\begin{table}[b!]
	\caption{Fraction of zeros (\%). Notation follows Table~3 of the main text.}
	\label{tab:SparsenessMainTable}
	\centering
	\begin{tabular}{>{\centering}p{0.15\textwidth}>{\centering}p{0.01\textwidth}>{\centering}p{0.15\textwidth}>{\centering}p{0.15\textwidth}>{\centering\arraybackslash}p{0.3\textwidth}}
		\Xhline{1.2pt}
	    \multicolumn{1}{l}{\multirow{2}{*}{Model \& Dataset}} & & \multicolumn{3}{c}{Fraction of zeros (\%)} \\ \cline{3-5}
        & &  LSQ & QLS (Ours) & QLS $\rightarrow$ 1\% LSQ \\ \hline
		 \multicolumn{1}{l}{$4$-bit ResNet-$18$ \& CIFAR-100} & & \multicolumn{1}{c}{60.71} & \multicolumn{1}{c}{74.13} & \multicolumn{1}{c}{74.09}\\\hline
		 \multicolumn{1}{l}{$4$-bit ResNet-$18$ \& ImageNet} & & \multicolumn{1}{c}{18.41} & \multicolumn{1}{c}{19.74} & \multicolumn{1}{c}{19.74}\\\hline
		 \multicolumn{1}{l}{$3$-bit RFDN~\citeapp{liu2020residual_app} \& DIV2K~\citeapp{agustsson2017ntire_app}} & & \multicolumn{1}{c}{9.64} & \multicolumn{1}{c}{19.13} & \multicolumn{1}{c}{18.88}\\         
        \Xhline{1.2pt}
	\end{tabular}
\end{table}
\begin{figure}[b!]
    \includegraphics[width=1.0\linewidth]{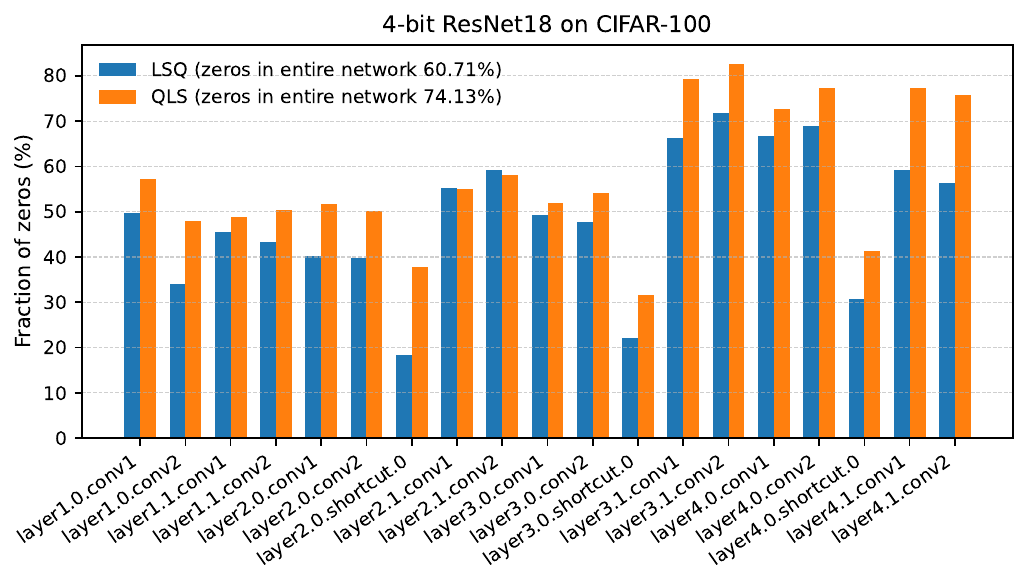}
    \caption{Fraction of zero weights across layers in W4Afp ResNet-18/CIFAR-100.}
    \label{fig:sparseness_resnet}
\end{figure}
\begin{figure}[htb]
    \includegraphics[width=1.0\linewidth]{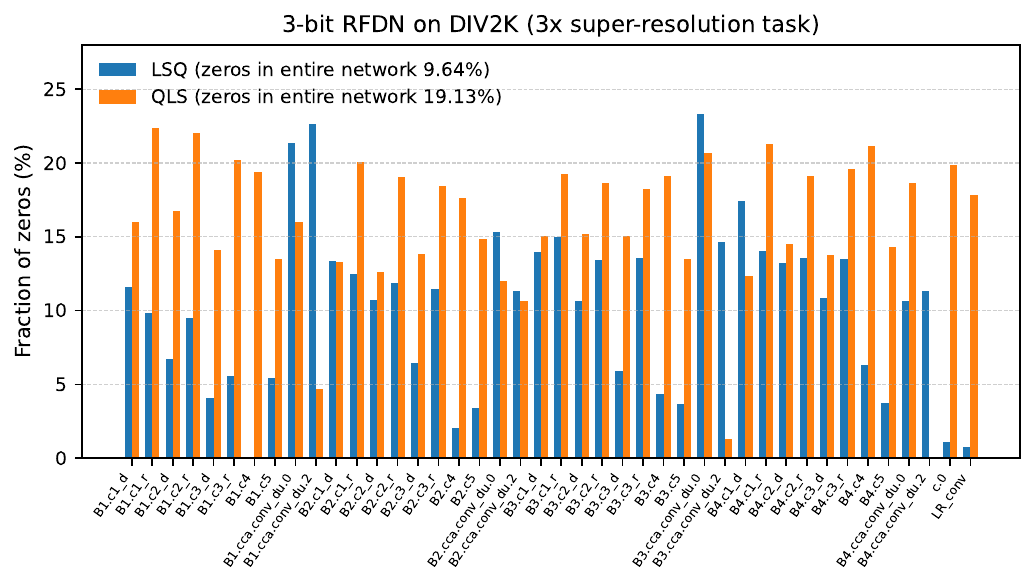}
    \caption{Fraction of zero weights across layers in W3Afp RFDN model.}
    \label{fig:sparseness_rfdn}
\end{figure}
During the QLS training, we observe that a substantial fraction of tensor elements corresponding to the endpoints $\mathbf{\Theta}^{(1)}$ and $\mathbf{\Theta}^{(2)}$ of the subspace converge to approximately symmetric values with respect to zero, \ie, a large fraction of tensor elements exhibit approximate antisymmetry $[\mathbf{\Theta}^{(1)}]_i \approx -[\mathbf{\Theta}^{(2)}]_i$. As a consequence, the midpoint of the learned subspace, $\mathbf{\Theta}^{(\mathrm{mid})}={(\mathbf{\Theta}^{(1)}+\mathbf{\Theta}^{(2)})}/{2},$
contains a large amount of near-zero parameters. After quantization, these small-magnitude weights are mapped exactly to zero, resulting in a significantly increased level of sparsity in the quantized model.

Table~\ref{tab:SparsenessMainTable} shows the percentage of zero-valued weights in the networks quantized by QLS/LSQ. Overall, QLS consistently produces solutions with higher sparsity. Notably, strong sparsity emerges in the RFDN super-resolution model despite the absence of explicit weight-magnitude regularization (e.g., weight decay). For image classification networks, the distribution of zero weights for LSQ and QLS exhibits moderate correlations across layers (see~\cref{fig:sparseness_resnet}). In contrast, for RFDN we observe a pronounced redistribution of zeros across layers under QLS training, leading to the sparsity pattern which differs substantially from one produced by LSQ (see~\cref{fig:sparseness_rfdn}).

Finally, we find that subsequent LSQ fine-tuning (\cref{tab:SparsenessMainTable}) of post-training QLS-quantized models does not significantly degrade the sparsity level, indicating that the sparse structure induced by QLS is relatively stable under further optimization.

\section{Additional Results on Cross-Bit Adaptability and Power-of-Two Quantization}
\label{sec:more_cross_bit_pow2}
In this supplementary section, we present additional results on cross-bit adaptation scenarios, in which a model trained for $x$-bit quantization scheme and then evaluated under a different inference precision $y$. Following the notation of the main paper, for the QLS method we denote this setting as QLS($x,y$) $\equiv$ QLS($x$-bit train, $y$-bit inference).

Table~\ref{tab:bit_inference} presents results for both regimes: (i) inference at a higher bitwidth than used during training ($y>x$, see the discussion in the main text), and (ii) inference at a lower bitwidth ($y<x$). In the latter case, Eq.~(10) of the main paper gives
\begin{equation}\label{eq:greater_scale}
\mathbf{s}^{(\mathrm{mid})}(b=y) \approx 2^{x-y}\times\mathbf{s}^{(\mathrm{mid})}(b=x)\in \{ 2, 4, 8, \dots \}\mathbf{s}^{(\mathrm{mid})}(b=x),
\end{equation}
indicating that the scale factor for $y$-bit ($y<x$) quantization derived from the subspace midpoint becomes a power-of-two multiple $(2^{n>0})$ of the scale corresponding to $x$-bit quantization. Consequently, some quantized values fall outside the trained low-loss subspace. Nevertheless, a fraction of the weights remains within the high-accuracy subspace even after quantization. As a result, although QLS($x,y<x$) exhibits a certain accuracy degradation, it is considerably less pronounced than that one observed for LSQ.

In Table~\ref{tab:power_of_two_inference}, we further examine the combined effect of the cross-bit adaptation and the transformation of scale factors to a hardware-friendly power-of-two representation, $\mathbf{s} \rightarrow \mathbf{s}_{\text{pow2}} = 2^{\lceil \log_2(\mathbf{s}) \rceil}$. In this scenario, the advantages of the proposed QLS scheme become even more pronounced.
\begin{table}[t]
	\centering
    \caption{Average accuracy (\%) over 3 runs for different train–inference bitwidth configurations. For LSQ, the best result across the two bitwidth-adaptation schemes (Sec.~4.2 of the paper) is presented.}
    \label{tab:bit_inference}
	\begin{tabular}{>{\centering}p{0.05\textwidth}>{\centering}p{0.005\textwidth}>{\centering}p{0.001\textwidth}>{\centering}p{0.05\textwidth}>{\centering}p{0.03\textwidth}>{\centering}p{0.08\textwidth}>{\centering}p{0.08\textwidth}>{\centering}p{0.08\textwidth}>{\centering}p{0.06\textwidth}>{\centering}p{0.08\textwidth}>{\centering}p{0.08\textwidth}>{\centering\arraybackslash}p{0.08\textwidth}}
		\Xhline{1.2pt}
	    \multirow{3}{*}{\rotatebox[origin=c]{90}{\parbox[c]{1.5cm}{\centering Model}}} & \multicolumn{1}{l}{\multirow{3}{*}{Dataset}} & & \multicolumn{1}{c}{\multirow{3}{*}{\begin{tabular}[c]{@{}c@{}c@{}} Bitwidth \\ during \\ \textbf{train} \end{tabular}}} & & \multicolumn{7}{c}{Bitwidth during \textbf{inference}} \\ \cline{6-12}
        & & & & & \multicolumn{3}{c}{LSQ} & & \multicolumn{3}{c}{QLS (Ours)} \\ \cline{6-8} \cline{10-12}
        & & & & & $4$-bit & $5$-bit & $6$-bit & & $4$-bit & $5$-bit & $6$-bit \\\hline
	\multirow{6}{*}{\rotatebox[origin=c]{90}{\parbox[c]{1.5cm}{\centering ResNet-$50$}}} & \multicolumn{1}{l}{\multirow{3}{*}{\begin{tabular}[l]{@{}l@{}}Tiny- \\ ImageNet\end{tabular}}} & & \multicolumn{1}{c}{$4$-bit} & & \multicolumn{1}{c}{\crosscell{a}{$67.47$}} & \multicolumn{1}{c}{$67.05$} & \multicolumn{1}{c}{$67.14$} & & \multicolumn{1}{c}{\crosscell{b}{$70.01$}} & \multicolumn{1}{c}{$70.19$} & \multicolumn{1}{c}{$70.19$}\\
    & & & \multicolumn{1}{c}{$5$-bit} & & \multicolumn{1}{c}{$67.99$} & \multicolumn{1}{c}{$69.14$} & \multicolumn{1}{c}{$68.69$} & & \multicolumn{1}{c}{$69.58$} & \multicolumn{1}{c}{$70.14$} & \multicolumn{1}{c}{$70.19$}\\
    & & & \multicolumn{1}{c}{$6$-bit} & & \multicolumn{1}{c}{$66.91$} & \multicolumn{1}{c}{$69.61$} & \multicolumn{1}{c}{$70.09$} & & \multicolumn{1}{c}{$69.55$} & \multicolumn{1}{c}{$70.08$} & \multicolumn{1}{c}{$70.30$}\\ \cline{2-12}   
	& \multicolumn{1}{l}{\multirow{3}{*}{ImageNet}} & & \multicolumn{1}{c}{$4$-bit} & & \multicolumn{1}{c}{\crosscell{c}{$76.10$}} & \multicolumn{1}{c}{$75.0\phantom{0}$} & \multicolumn{1}{c}{$75.0\phantom{0}$} & & \multicolumn{1}{c}{\crosscell{d}{$75.85$}} & \multicolumn{1}{c}{$76.32$} & \multicolumn{1}{c}{$76.41$}\\ 
    & & & \multicolumn{1}{c}{$5$-bit} & & \multicolumn{1}{c}{$71.02$} & \multicolumn{1}{c}{$76.41$} & \multicolumn{1}{c}{$75.76$} & & \multicolumn{1}{c}{$74.79$} & \multicolumn{1}{c}{$76.56$} & \multicolumn{1}{c}{$76.72$}\\
    & & & \multicolumn{1}{c}{$6$-bit} & & \multicolumn{1}{c}{$69.15$} & \multicolumn{1}{c}{$75.49$} & \multicolumn{1}{c}{$76.68$} & & \multicolumn{1}{c}{$73.92$} & \multicolumn{1}{c}{$76.04$} & \multicolumn{1}{c}{$76.82$}\\    
    \Xhline{1.2pt}

    
	\end{tabular}
\end{table}

\begin{table}[t]
	\centering
    \caption{The same as in Table~\ref{tab:bit_inference}, but with quantization scale factors converted directly to a power-of-two representation.}
    \label{tab:power_of_two_inference}
	\begin{tabular}{>{\centering}p{0.05\textwidth}>{\centering}p{0.005\textwidth}>{\centering}p{0.001\textwidth}>{\centering}p{0.05\textwidth}>{\centering}p{0.03\textwidth}>{\centering}p{0.08\textwidth}>{\centering}p{0.08\textwidth}>{\centering}p{0.08\textwidth}>{\centering}p{0.06\textwidth}>{\centering}p{0.08\textwidth}>{\centering}p{0.08\textwidth}>{\centering\arraybackslash}p{0.08\textwidth}}
		\Xhline{1.2pt}
	    \multirow{3}{*}{\rotatebox[origin=c]{90}{\parbox[c]{1.5cm}{\centering Model}}} & \multicolumn{1}{l}{\multirow{3}{*}{Dataset}} & & \multicolumn{1}{c}{\multirow{3}{*}{\begin{tabular}[c]{@{}c@{}c@{}} Bitwidth \\ during \\ \textbf{train} \end{tabular}}} & & \multicolumn{7}{c}{Bitwidth during \textbf{power-of-2 inference}} \\ \cline{6-12}
        & & & & & \multicolumn{3}{c}{LSQ} & & \multicolumn{3}{c}{QLS (Ours)} \\ \cline{6-8} \cline{10-12}
        & & & & & $4$-bit & $5$-bit & $6$-bit & & $4$-bit & $5$-bit & $6$-bit \\\hline
	\multirow{6}{*}{\rotatebox[origin=c]{90}{\parbox[c]{1.5cm}{\centering ResNet-$50$}}} & \multicolumn{1}{l}{\multirow{3}{*}{\begin{tabular}[l]{@{}l@{}}Tiny- \\ ImageNet\end{tabular}}} & & \multicolumn{1}{c}{$4$-bit} & & \multicolumn{1}{c}{$62.6\phantom{0}$} & \multicolumn{1}{c}{$63.77$} & \multicolumn{1}{c}{$62.30$} & & \multicolumn{1}{c}{$69.11$} & \multicolumn{1}{c}{$70.10$} & \multicolumn{1}{c}{$70.13$}\\
    & & & \multicolumn{1}{c}{$5$-bit} & & \multicolumn{1}{c}{$62.44$} & \multicolumn{1}{c}{$65.58$} & \multicolumn{1}{c}{$66.48$} & & \multicolumn{1}{c}{$69.28$} & \multicolumn{1}{c}{$69.98$} & \multicolumn{1}{c}{$70.07$}\\
    & & & \multicolumn{1}{c}{$6$-bit} & & \multicolumn{1}{c}{$64.5\phantom{0}$} & \multicolumn{1}{c}{$68.0\phantom{0}$} & \multicolumn{1}{c}{$68.02$} & & \multicolumn{1}{c}{$69.01$} & \multicolumn{1}{c}{$70.1\phantom{0}$} & \multicolumn{1}{c}{$70.32$}\\ \cline{2-12}   
	& \multicolumn{1}{l}{\multirow{3}{*}{ImageNet}} & & \multicolumn{1}{c}{$4$-bit} & & \multicolumn{1}{c}{$72.45$} & \multicolumn{1}{c}{$72.45$} & \multicolumn{1}{c}{$72.45$} & & \multicolumn{1}{c}{$73.13$} & \multicolumn{1}{c}{$75.97$} & \multicolumn{1}{c}{$76.34$}\\ 
    & & & \multicolumn{1}{c}{$5$-bit} & & \multicolumn{1}{c}{$58.34$} & \multicolumn{1}{c}{$73.98$} & \multicolumn{1}{c}{$75.03$} & & \multicolumn{1}{c}{$73.4\phantom{0}$} & \multicolumn{1}{c}{$75.93$} & \multicolumn{1}{c}{$76.55$}\\
    & & & \multicolumn{1}{c}{$6$-bit} & & \multicolumn{1}{c}{$53.19$} & \multicolumn{1}{c}{$74.45$} & \multicolumn{1}{c}{$75.87$} & & \multicolumn{1}{c}{$61.65$} & \multicolumn{1}{c}{$75.97$} & \multicolumn{1}{c}{$76.61$}\\    
    \Xhline{1.2pt}    
	\end{tabular}
\end{table}    
\section{Additional Results on Full Quantization}
As discussed in the main paper (see also the~\cref{sec:more_cross_bit_pow2} of this supplemental material), a model trained with QLS for a given weight bitwidth can be directly adapted to higher bitwidths without additional fine-tuning, resulting in a monotonic improvement in performance. Furthermore, the midpoint of the learned subspace provides a strong initialization for subsequent full quantization of both weights and activations (Sec.~4.2 of the paper).

Here we demonstrate that this property eliminates the need to retrain the subspace for each target full-quantization configuration. In Table~6, we report results where both weights and activations are quantized to $b$ bits (W$b$A$b$). We compare two strategies: (i) \textbf{Conventional LSQ} -- starting from an FP model and applying LSQ to both weights and activations; (ii) \textbf{Combined QLS–LSQ} -- initializing from the midpoint of the learned subspace, applying QLS to the weights (without explicit quantizers or the STE approximation during training), and LSQ to the activations.

Importantly, the second strategy uses the same midpoint initialization for all target bitwidths. As shown in~\cref{{tab:WbAb}}, the combined QLS–LSQ approach consistently achieves faster convergence and improved accuracy, both under a single training epoch and when trained with the full training pipeline of FP model.

\label{sec:more_full_quantization}
\begin{table}[t]
\caption{Average PSNR of the WbAb quantized RFDN model (notation as in Table~4 of the main paper).}
\label{tab:WbAb}
\centering
\begin{tabular}{l @{\hspace{5pt}} l c l @{\hspace{12pt}} c @{\hspace{12pt}} c @{\hspace{12pt}} c}
\Xhline{1.2pt}
\multicolumn{4}{l}{Quantization Setup} & Epochs & Set5 & Urban100\\
\hline
\multirow{4}{*}{\rotatebox[origin=c]{90}{\parbox[c]{1.5cm}{\centering W4A4}}} & Standard FP & $\rightarrow$ & $\text{W4}_{\text{LSQ}}\text{A4}_{\text{LSQ}}$ & \multirow{2}{*}{1} & 22.05 & 20.37 \\
& $\mathbf{\Theta}^{({\rm mid})}(b=4)$ & $\rightarrow$ & $\text{W4}_{\textbf{QLS}}\text{A4}_{\text{LSQ}}$ &  & \textcolor{red}{31.0\phantom{0}} & \textcolor{red}{25.68} \\\cline{5-7}
& Standard FP & $\rightarrow$ & $\text{W4}_{\text{LSQ}}\text{A4}_{\text{LSQ}}$ & \multirow{2}{*}{1000} & 31.01 & 24.89 \\
& $\mathbf{\Theta}^{({\rm mid})}(b=4)$ & $\rightarrow$ & $\text{W4}_{\textbf{QLS}}\text{A4}_{\text{LSQ}}$ &  & \textcolor{red}{32.85} & \textcolor{red}{26.29} \\\hline
\multirow{4}{*}{\rotatebox[origin=c]{90}{\parbox[c]{1.5cm}{\centering W5A5}}} & Standard FP & $\rightarrow$ & $\text{W5}_{\text{LSQ}}\text{A5}_{\text{LSQ}}$ & \multirow{2}{*}{1} & 33.43 & 26.97 \\
& $\mathbf{\Theta}^{({\rm mid})}(b=4)$ & $\rightarrow$ & $\text{W5}_{\textbf{QLS}}\text{A5}_{\text{LSQ}}$ &  & \textcolor{red}{33.7\phantom{0}} & \textcolor{red}{27.15} \\\cline{5-7}
& Standard FP & $\rightarrow$ & $\text{W5}_{\text{LSQ}}\text{A5}_{\text{LSQ}}$ & \multirow{2}{*}{1000} & 34.02 & 27.66 \\
& $\mathbf{\Theta}^{({\rm mid})}(b=4)$ & $\rightarrow$ & $\text{W5}_{\textbf{QLS}}\text{A5}_{\text{LSQ}}$ &  & \textcolor{red}{34.12} & \textcolor{red}{27.7\phantom{0}} \\\hline
\multirow{4}{*}{\rotatebox[origin=c]{90}{\parbox[c]{1.5cm}{\centering W6A6}}} & Standard FP & $\rightarrow$ & $\text{W6}_{\text{LSQ}}\text{A6}_{\text{LSQ}}$ & \multirow{2}{*}{1} & 33.83 & 27.29 \\
& $\mathbf{\Theta}^{({\rm mid})}(b=4)$ & $\rightarrow$ & $\text{W6}_{\textbf{QLS}}\text{A6}_{\text{LSQ}}$ &  & \textcolor{red}{33.95} & \textcolor{red}{27.31} \\\cline{5-7}
& Standard FP & $\rightarrow$ & $\text{W6}_{\text{LSQ}}\text{A6}_{\text{LSQ}}$ & \multirow{2}{*}{1000} & 34.08 & 27.76 \\
& $\mathbf{\Theta}^{({\rm mid})}(b=4)$ & $\rightarrow$ & $\text{W6}_{\textbf{QLS}}\text{A6}_{\text{LSQ}}$ &  & \textcolor{red}{34.17} & \textcolor{red}{27.8\phantom{0}} \\
\Xhline{1.2pt}
\end{tabular}
\end{table}
\section{Limitations and Future Work}
\label{sec:limits_next}
The proposed QLS algorithm finds low-loss regions of the objective landscape aligned with a target quantization scheme. The QLS quantization scheme learns these quantization-aware subspaces during training from scratch, which requires reproducing the full FP training pipeline. However, this process often produces improved FP models compared to standard training.

In principle, the subspace could also be learned from a pretrained FP model. Preliminary experiments suggest this may require far fewer training iterations, though appropriate learning-rate adjustments remain unclear and appear to be task-dependent.

Extending the QLS method to large-scale transformer models for vision and language is an important direction for future work. Because training such models from scratch is computationally expensive, we leave this to a separate study that may also require further adaptations to the QLS pipeline.

Finally, the current work focuses primarily on weight-only quantization. Extending QLS to activation quantization is feasible and will be explored in future work.

\clearpage
\begin{algorithm}[t]
\caption{Pseudo code of QLSConv2d layer}
\label{alg:qlsconv}
\begin{algorithmic}[1]

\Require Input feature map $x$, Conv2D layer $m$, bit-width $b$
\Ensure Output feature map $y$

\State Initialize learnable weights $W_1, W_2 \leftarrow \text{zeros\_like}(m.weight)$
\State Initialize interpolation tensor $\alpha \leftarrow 0.5 \, \text{ones\_like}(m.weight)$  
\If{$m.bias$ exists}
    \State $bias \leftarrow m.bias$
\EndIf

\State $t_{neg} \leftarrow -2^{b-1}+1$
\State $t_{pos} \leftarrow 2^{b-1}-1$

\State $W_1 \leftarrow$ KaimingNormal()
\State $W_2 \leftarrow$ KaimingNormal()

\vspace{2mm}

\Function{QuantizeMidpoint}{$W_{mid}$}
    \State $scale \gets \frac{2 \cdot \max(|W_{mid}|)}{t_{pos}-t_{neg}}$
    \State $Q \leftarrow \text{clamp}(\text{round}(W_{mid}/scale), t_{neg}, t_{pos})$
    \State \Return $scale \cdot Q$
\EndFunction

\vspace{2mm}

\Function{ForwardTrain}{$x$}
    \State $W \leftarrow (1-\alpha)W_1 + \alpha W_2$
    \State \Return $\text{Conv2D}(x, W, bias)$
\EndFunction

\vspace{2mm}

\Function{ForwardTest}{$x$}
    \State $W_{mid} \leftarrow (W_1 + W_2)/2$
    \State $W_q \leftarrow \textsc{QuantizeMidpoint}(W_{mid})$
    \State \Return $\text{Conv2D}(x, W_q, bias)$
\EndFunction

\vspace{2mm}

\Function{Forward}{$x$, training}
    \If{training}
        \State \Return \Call{ForwardTrain}{$x$}
    \Else
        \State \Return \Call{ForwardTest}{$x$}
    \EndIf
\EndFunction

\end{algorithmic}
\end{algorithm}

\begin{algorithm}[t]
\caption{Pseudo code of $\mathcal{R}_{\rm qdist}$ regularization calculation}\label{alg:qdist}
\begin{algorithmic}[1]

\Require Model with QLSConv2d layers
\Ensure Subspace regularization loss $R_{qdist}$

\State $R_{qdist} \gets 0$
\State $n_{QLS} \gets 0$

\ForAll{layers $m$ in model.modules()}
    \If{$m$ is instance of QLSConv2d}
        \State $W_{mid} \gets (m.W_1 + m.W_2)/2$
        \State $scale \gets \frac{2 \cdot \max(|W_{mid}|)}{m.t_{pos}-m.t_{neg}}$
        \State $W_{norm} \gets |m.W_1 - m.W_2| / scale$
        \State $obj \gets \sum \textbf{where}(W_{norm} < 1, (W_{norm} - 1) ^ 2, 0)$
        \State $R_{qdist} \gets R_{qdist} + obj / \text{numel}(W_{mid})$
        \State $n_{QLS} \gets n_{QLS} + 1$
    \EndIf
\EndFor

\State \Return $R_{qdist} / n_{QLS}$

\end{algorithmic}
\end{algorithm}

\begin{algorithm}[t]
\caption{Training Step for model with QLSConv2d Layers}
\label{alg:train_classification}
\begin{algorithmic}[1]

\Require Model with QLSConv2d layers, training loader, optimizer, scheduler, criterion, regularization weight $\lambda$
\Ensure Updated model parameters

\State Set model to \textbf{training mode}

\ForAll{batches $(x, y)$ in training loader}
    \ForAll{layers $m$ in model.modules()}
        \If{$m$ is instance of QLSConv2d}
            \State $m.\alpha \gets$ random tensor of the same shape as $m.\alpha$ from a uniform distribution on the interval $[0,1]^{m.\alpha.{\rm shape}}$ 
        \EndIf
    \EndFor
    
    \If{scheduler exists}
        \State scheduler.step()
    \EndIf
    
    \State optimizer.zero\_grad()
    \State $\hat{y} \gets$ model$(x)$
    
    \State $L_{task} \gets$ criterion$(\hat{y}, y)$
    \State $R_{qdist} \gets$ qdist\_regularization(model) \Comment{See~\cref{alg:qdist}} 
    \State $L_{total} \gets L_{task} + \lambda \cdot R_{qdist}$
    
    \State Backpropagate $L_{total}$
    \State optimizer.step()
\EndFor

\end{algorithmic}
\end{algorithm}

\clearpage
\bibliographystyleapp{splncs04}
\bibliographyapp{SM}
\end{document}